%% file: acl_latex.tex
\pdfoutput=1
\PassOptionsToPackage{dvipsnames}{xcolor}
\documentclass[11pt]{article}

\usepackage{acl}

\usepackage{times}
\usepackage{latexsym}

\usepackage[T1]{fontenc}

\usepackage[utf8]{inputenc}

\usepackage{microtype}

\usepackage{inconsolata}

\usepackage{booktabs}
\usepackage{multirow}
\usepackage{microtype}
\usepackage{amsmath}
\usepackage{amssymb}
\usepackage{amsfonts}
\usepackage{bbm}
\usepackage{diagbox}
\usepackage{graphicx}
\usepackage{makecell}
\usepackage{xspace}
\usepackage{mathabx}
\usepackage[dvipsnames]{xcolor}
\usepackage[normalem]{ulem}
\usepackage{caption,subcaption}
\usepackage{stackengine}

\newcommand{\tabref}[1]{Tab.~\ref{tab:#1}}
\newcommand{\figref}[1]{Fig.~\ref{fig:#1}}
\newcommand{\appref}[1]{App.~\ref{#1}}

\newcommand{\subsecref}[1]{§\ref{subsec:#1}}

\newcommand{\superni}{\textsc{Super-NaturalInstructions}\xspace}

\newcommand{\bias}{\textsl{BiasScore}\xspace}
\newcommand{\rsd}{\textsl{RSD}\xspace}

\title{Beyond Performance:\\ Quantifying and Mitigating Label Bias in LLMs}

\author{Yuval Reif \quad\quad Roy Schwartz\\
  School of Computer Science and Engineering, The Hebrew University of Jerusalem, Israel \\
  \texttt{\{yuval.reif,roy.schwartz1\}@mail.huji.ac.il}}

\begin{document}
\maketitle

\input{text/0.abstract}
\input{text/1.introduction}

\input{text/2.metrics}

\input{text/3.setting}

\input{text/4.results}

\input{text/5.calibration}
\input{text/6.analysis}

\input{text/_1.related_work}
\input{text/_1.conclusion.tex}

\input{text/_1.limitations.tex}

\input{text/_1.acknowledgements.tex}

\bibliography{anthology,custom}

\input{text/_1.appendix}

\end{document}

%% file: text/0.abstract.tex
\begin{abstract}
Large language models (LLMs) have shown remarkable adaptability to diverse tasks, by leveraging context prompts containing instructions, or minimal input-output examples. However, recent work revealed they also exhibit \emph{label bias}---an undesirable preference toward predicting certain answers over others. Still, detecting and measuring this bias reliably and at scale has remained relatively unexplored. In this study, we evaluate different approaches to quantifying \emph{label bias} in a model's predictions, conducting a comprehensive investigation across 279 classification tasks and ten LLMs. Our investigation reveals substantial label bias in models both before and after debiasing attempts, as well as highlights the importance of outcomes-based evaluation metrics, which were not previously used in this regard. We further propose a novel label bias calibration method tailored for few-shot prompting, which outperforms recent calibration approaches for both improving performance and mitigating label bias. Our results emphasize that label bias in the predictions of LLMs remains a barrier to their reliability.\footnote{We release our code at \url{https://github.com/schwartz-lab-NLP/label-bias}.}

\end{abstract}

%% file: text/1.introduction.tex
\section{Introduction}

\begin{figure}[!t]
    \centering
    \includegraphics[width=1\linewidth]{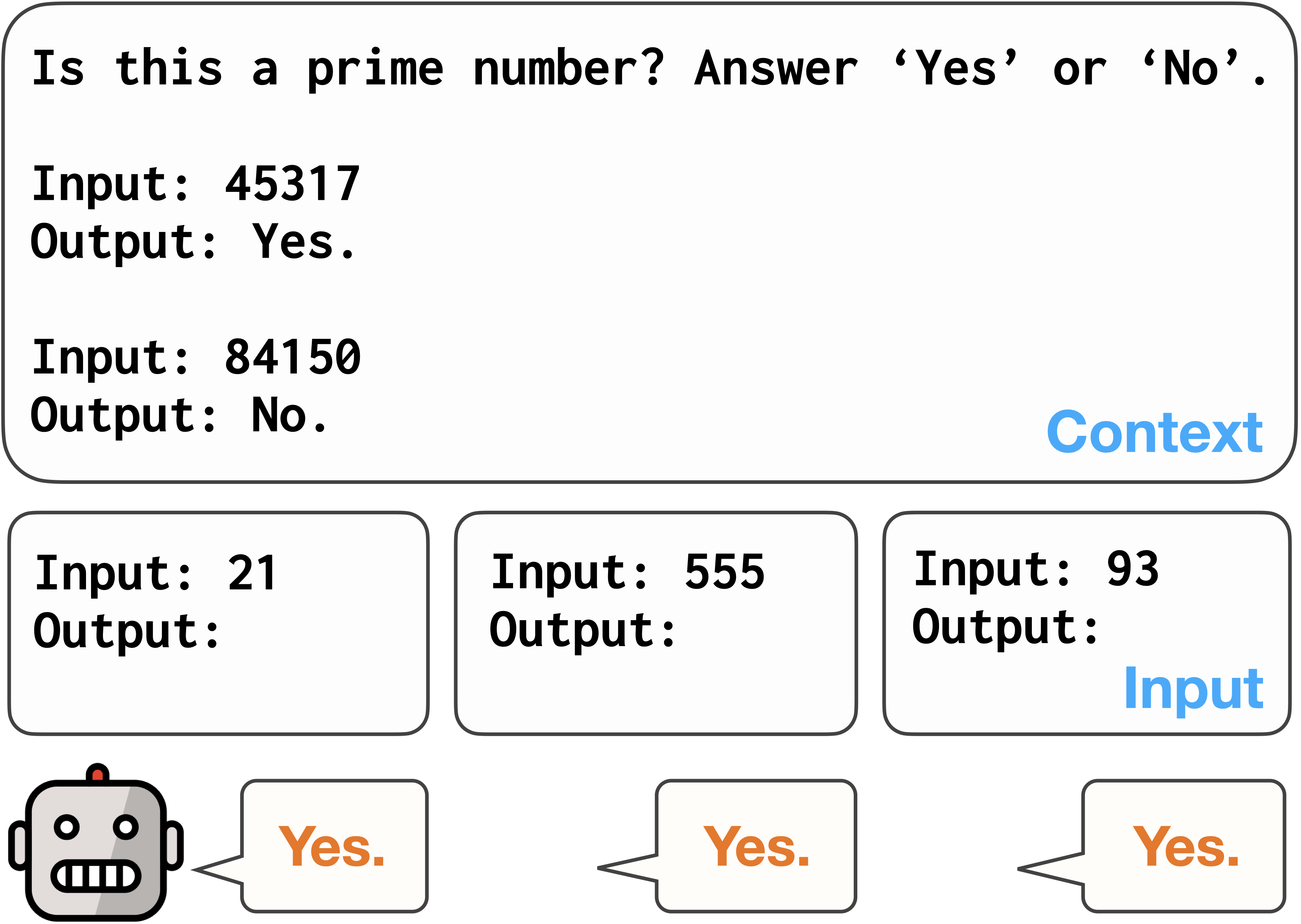}
    \caption{LLMs exhibit \textit{label bias}---a tendency to output a given label regardless of the context and input (in this example, `yes' over `no'). In this work we evaluate LLM label bias across ten LLMs and 279 classification tasks, showing label bias is a major problem in LLMs.}
    \label{fig:fig1}
\end{figure}

Large language models (LLMs) have demonstrated impressive abilities in adapting to new tasks when conditioned on a context prompt, containing task-solving instructions \cite{wei2022finetuned} or few examples of input-output pairs \cite{brown2020language}. 
Still, recent work has shown that predictions of LLMs exhibit \emph{label bias}---a strong, undesirable preference towards predicting certain answers over others~(\citealp{zhao2021calibrate, chen2022relation,fei-etal-2023-mitigating}, see \figref{fig1}).
Such preferences were shown to be affected by the choice and order of in-context demonstrations \cite{liu-etal-2022-makes,lu-etal-2022-fantastically}, the model's pretraining data \cite{dong-etal-2022-calibrating}, or textual features of the task data \cite{fei-etal-2023-mitigating}. Consequently, several approaches were proposed to address this problem, mostly by calibrating the model's output probabilities to compensate for this bias \cite{zhao2021calibrate,fei-etal-2023-mitigating}. 

Despite these efforts, label bias evaluation relies on \emph{performance} metrics such as accuracy, rather than metrics designed to directly quantify the \emph{bias}.
In doing so, we might inadvertently overlook crucial aspects of model behavior.
Indeed, although a given method could effectively improve performance, substantial bias might still persist in the model's predictions---deeming the method insufficient and the model unreliable.
Alternatively, performance could remain relatively unchanged, but with the bias mostly removed.

In this work, we take a step towards a more comprehensive understanding of the extent of label bias in LLMs and the effects of mitigation approaches.
Using metrics to directly measure the label bias in model predictions, which we derive from previous work on fairness and label bias estimation, we evaluate ten LLMs on 279 diverse classification and multiple-choice tasks from \superni~\cite{wang-etal-2022-super}. We examine both performance and bias along axes such as scale and number of in-context demonstrations. 
We also evaluate the impact of label bias mitigation methods, such as calibration and few-shot LoRA fine-tuning~\cite{hu2022lora}. 

Our investigation reveals substantial label bias in the predictions of LLMs across all evaluated settings, indicating that raw LLM output scores often represent simple, heuristic solutions. While increasing model size, providing in-context demonstrations, and instruction-tuning all contribute to reducing bias, ample bias persists, even after applying mitigation methods.
Surprisingly, these results also hold for tasks where the labels are all semantically equivalent (e.g., in multi-choice question answering). 
Further, although the examined calibration methods can reduce bias and improve performance, we also find cases where they negatively impact both bias and overall performance.

Motivated by these findings, we propose a novel calibration method for few-shot prompting
that more accurately estimates a model's label bias, using only its predictions on the in-context demonstrations. Compared to existing LLM bias calibration methods, our method improves performance while also removing considerably more bias. 

Our findings highlight the necessity of considering and measuring biases in the predictions of LLMs when evaluating their performance. Moreover, adjusting models to their tasks through more accurate and effective estimation of biases holds promise for improving the reliability of LLMs and their applications.

%% file: text/2.metrics.tex
\section{LLM Label Bias}
Our objective is to broaden the understanding of label bias in LLMs and the effectiveness of mitigation strategies, focusing on classification tasks.
In this section, we define metrics designed to quantify bias in model predictions, providing a nuanced examination of label bias that extends beyond traditional performance metrics.
We describe the setting of label bias in in-context learning~(\subsecref{background_notations}), briefly outline methods to mitigate it~(\subsecref{background_mitigation}), and finally review approaches to evaluate label bias as well as define the metrics we use in this work~(\subsecref{metrics}). 

\subsection{Label Bias}
\label{subsec:background_notations}
When employing LLMs for classification tasks through prompting, the model is given a test example $x$, preceded by a context $C$. This context can contain a (potentially empty) set of examples of the task's input-output mapping~$[(x^1, y^1), \ldots , (x^k, y^k)]$, henceforth \textit{demonstrations}, and may also include task instructions. To determine the model's prediction from a set of answer choices $Y$, the likelihood it assigns to each continuation $y \in Y$ is computed, and the highest probability option is taken as the model prediction:\begin{equation*}
    \arg\max_{y \in Y} p(y \mid x, C)
\end{equation*}
These output probabilities often exhibit \emph{label bias}, where the model tends to assign higher probability to certain answers regardless of the input test example $x$ (\figref{fig1}).
Multiple factors were posited to influence this bias, including the choice of verbalizers $Y$, the choice and order of in-context examples in $C$, and the overall textual features of task input~$x$~\cite{zhao2021calibrate,fei-etal-2023-mitigating}.

\subsection{Bias Mitigation}

\label{subsec:background_mitigation}
The predominant approach to alleviate label bias is to calibrate the model's output probabilities post-hoc, for a specific context prompt $C$. 
Such methods typically first estimate the model's label bias using its output probabilities on a set of inputs, which can be content-free (e.g., ``N/A'' or random words from the task's domain; \citealt{zhao2021calibrate,fei-etal-2023-mitigating}) or ordinary task inputs \cite{han2023prototypical}. Next, calibration parameters are chosen based on this estimate, and used to adjust the original output probabilities during inference to generate the (hopefully unbiased) output.

\subsection{Evaluation Measures}
\label{subsec:metrics}
Most LLM label bias analysis relies on indirect assessments. For instance, some work inspected improvements in overall performance gained after applying techniques to mitigate it~\cite{fei-etal-2023-mitigating, holtzman-etal-2021-surface, zhao2021calibrate}. However, these do not indicate the extent of bias originally present, or that which remains after mitigation.
We next examine approaches to measure this bias more directly, and define the metrics we use in this work.
Importantly, we focus on label bias measures that could be used effectively both \textit{before} and \textit{after} applying mitigation techniques such as calibration.

Drawing from previous research on fairness and bias in machine learning, we observe that there are two distinct yet related aspects in which label bias can be measured in LLM predictions: through the probabilities assigned by the model to different answers, e.g., assigning the label ``yes'' with an \textit{average} output probability of 0.55, while ``no'' with 0.45; and through the model's predictions for different labels, e.g., achieving a recall of 0.50 for instances labeled ``yes'', compared to 0.40 on ``no''~\cite{mehrabi2021survey}. 
Below we describe methods to measure each of these notions of bias.

\paragraph{Probabilistic approach} Previous work used qualitative assessments to visualize model output distributions on selected datasets~\cite{zhao2021calibrate, han2023prototypical}. However, these cannot be used to rigorously evaluate models at larger scales.
Recently, \citet{fei-etal-2023-mitigating} proposed to measure a model's label bias by considering two sets of inputs: a set of synthetic, content-free task inputs $\hat{X}_\textsl{cf}$, and inputs consisting of random vocabulary words $\hat{X}_\textsl{rand}$. For each input, they compute the output probabilities on every label $y\in Y$, and finally compute the model's \emph{mean} predicted probabilities across both sets,  $\hat{p}_\textsl{cf}$ and $\hat{p}_\textsl{rand}$:
\begin{equation*}
    \hat{p}_\ast(y) = \frac{1}{\lvert \hat{X}_\ast \rvert} \sum_{x\in \hat{X}_\ast} p(y \mid x, C)
\end{equation*}
The model's bias is then defined to be the total variation distance $d_{TV}$ between the two distributions:
\begin{equation*}
    d_{TV}(\hat{p}_\textsl{cf}, \hat{p}_\textsl{rand}) = \frac{1}{2} \sum_{y\in Y}\; \lvert\; \hat{p}_\textsl{cf}(y) - \hat{p}_\textsl{rand}(y) \;\rvert
\end{equation*}
Importantly, since \citet{fei-etal-2023-mitigating} also use the model's predictions on the content-free inputs $\hat{X}_\textsl{cf}$ to calibrate it, this metric cannot be used to quantify the label bias remaining after calibration. 

In this work, we simplify the computation of this metric and adapt it to be used after calibration.
First, we hold-out a set of inputs to be used exclusively for measuring bias. 
Second, when estimating the model's average output probabilities, instead of using synthetic inputs, we use in-distribution examples held-out from the test set, $\hat{X}_\textsl{i.d.} = \left((x_1, y_1), \ldots, (x_m, y_m)\right)$. %
This setup allows to account for label imbalance in the data used for bias estimation $\hat{X}_\textsl{i.d.}$, as the instances in the test set are all labeled. To do so, we first estimate the model's output distribution individually on each subset of examples with gold label $\ell\in Y$, $\hat{X}_\textsl{i.d.}^\ell = \{ (x,y) \in \hat{X}_\textsl{i.d.} \mid y = \ell\}$, by computing:
\begin{equation*}
    \hat{p}_\textsl{i.d.}^\ell (y) = \frac{1}{\lvert \hat{X}_\textsl{i.d.}^\ell \rvert} \sum_{x\in \hat{X}_\textsl{i.d.}^\ell} p(y \mid x, C)
\end{equation*}
and then set $\hat{p}_\textsl{i.d.}$ to be the average of these estimates.\footnote{In case examples for an infrequent label $\ell\in Y$ are not found in $\hat{X}_{i.d.}$, we exclude it from the computation of $\hat{p}_\textsl{i.d.}$.} 
Instead of $\hat{p}_\textsl{rand}$, we use the uniform distribution over all answer choices $(\frac{1}{|Y|}, \ldots, \frac{1}{|Y|})$, which recent mitigation approaches considered as the ``ideal'' and unbiased mean output distribution~\cite{zhao2021calibrate}. 
Finally, we define the model's \textbf{bias score} as the total variation distance between these two distributions:
\begin{equation*}
    BiasScore = \frac{1}{2} \sum_{y\in Y} \biggr\lvert \; \hat{p}_\textsl{i.d.}(y) - \frac{1}{|Y|} \; \biggl\rvert
\end{equation*}

\paragraph{Outcomes-based approach}
When considering the effects of label bias on model predictions, strong label bias will likely result in disparities in task performance on instances of different classes. However, metrics to assess such disparities were not used in previous analyses of label bias. 

We propose to use the \textbf{Relative Standard Deviation of class-wise accuracy} (\rsd; \citealt{croce2021robustbench, benz2021robustness}), 
a metric used for studying fairness in classification.
\rsd is defined as the standard deviation of the model's class-wise accuracy $(\textsl{acc}_1, \ldots, \textsl{acc}_{|Y|})$, divided by its mean accuracy $\textsl{acc}$ on the entire evaluation data:\footnote{The goal of this normalization is to enhance the metric's interpretability across tasks of varying difficulty.}
\begin{equation*}
    RSD = \frac{\sqrt{\frac{1}{|Y|}\sum_{i=1}^{|Y|}(\textsl{acc}_i - \textsl{acc})^2}}{\textsl{acc}}
\end{equation*}
Intuitively, \rsd is low when model performance~is similar on all classes, and high when it performs well on some classes but poorly on others.

\paragraph{Discussion}
We note that each evaluation approach could detect biases that the other does not. For example, a slight bias in the model's average output probabilities (e.g., 55\% vs.~45\%) could render dramatic bias in actual outcomes if the model \emph{always} assigns higher probability to some label. Conversely, when the output probabilities are biased \emph{on average} but the model's class-wise performance is balanced, this \emph{hidden} bias could result in actual performance disparities on more difficult~instances. We therefore suggest reporting both measures.

%% file: text/3.setting.tex
\section{Experimental Setting}
\label{sec:experimental_setting}

\subsection{Datasets}\label{subsec:datasets} We evaluate models on 279 diverse tasks from the \superni benchmark~\cite{wang-etal-2022-super}. We select all available classification and multi-choice question answering tasks where the output space is a set of predefined labels, such as ``yes/no'' or ``A/B/C''.
We sample 1,000 evaluation examples for all tasks with larger data sizes, and additionally sample 32 held-out examples for computing the bias score metric~(\subsecref{metrics}), and 64 more examples to use as a pool of instances for choosing in-context demonstrations and LoRA fine-tuning examples.
We only include tasks with at least 300 evaluation examples in our experiments. For details on the selected tasks, see \appref{app_superni}.

\subsection{Models and Evaluation Setup} 
We experiment with models of different sizes from three LLM families: Llama-2 7B and 13B~\cite{touvron2023llama}, Mistral 7B~\cite{jiang2023mistral}, and Falcon 7B and 40B~\cite{penedo2023falcon}. We use both the base and instruction fine-tuned versions of each model. We evaluate models using context prompts with $k \in \{0, 2, 4, 8, 16\}$ demonstrations, and average the results across $3$ different sets of demonstrations for each $k$. 
To control the evaluation budget, we run the more expensive Falcon 40B experiments with $k\in\{0, 8, 16\}$ averaged across $2$ sets of demonstrations.
We use the task instructions and prompt template defined in \superni.
For tasks where the answer choices $y\in Y$ have unequal token lengths, we use length-normalized log-likelihood to compute the output probabilities~\cite{holtzman-etal-2021-surface}.
For additional implementation details, see \appref{app_experimental_setting}.

\paragraph{Data contamination}
During their instruction tuning, Llama-2 chat models were initially fine-tuned on the \textsl{Flan} data collection~\cite{chung2022scaling, longpre2023flan}. 
As roughly 20\% of \textsl{Flan} consists of examples from \superni, our evaluation of Llama-2 instruction-tuned models is likely affected by data contamination~\cite{magar-schwartz-2022-data}. Still, our results show both 7B and 13B chat models exhibit extensive label bias, possibly due to later fine-tuning on other data. As it is unclear from the implementation details of~\citet{touvron2023llama} which exact instances in \superni were included in training, we do not take extra steps in attempt to reduce possible overlap and contamination. 

\subsection{Bias Mitigation Techniques}
\label{subsec:mitigation_methods}
We evaluate the effects of three label bias mitigation methods: two calibration methods designed to correct a model's label bias by adjusting its output scores; and few-shot LoRA fine-tuning~\cite{hu2022lora}, which adapts the model to the task and its label distribution.
We describe the methods below.

\label{subsec:calibration_methods}
\paragraph{Contextual calibration (CC)}
\citet{zhao2021calibrate} proposed to use calibration in order to remove the label bias arising from the context prompt $C$ and the model's pretraining. Inspired by confidence calibration methods~\cite{guo2017calibration}, they define a matrix $W$ that is applied to the model's original output probabilities $p$ during inference to obtain calibrated, debiased probabilities $q = \text{softmax}(Wp)$.
To determine the calibration parameters $W$, they first estimate the bias by  computing the model's average predicted probabilities $\hat{p}$ on a small set of ``placeholder'' content-free input strings, such as ``N/A'', which replace the ordinary task input that follows $C$.\footnote{As in the original implementation, we use ``N/A'', ``[MASK]'', and the empty string.}
Finally, they set $W = \text{diag}(\hat{p})^{-1}$, which ensures that the output class probabilities for the average content-free input are uniform, aiming to reduce bias on unseen examples.

\input{figures/before_mitigation/llama2}

\paragraph{Domain-context calibration (DC)}
Following CC,~\citet{fei-etal-2023-mitigating} proposed to estimate and mitigate the label bias arising from the textual distribution of the task's domain, by using task-specific content-free inputs to compute $\hat{p}$.
They construct such inputs by sampling and concatenating $L$ random words from the test set, where $L$ is the average instance input length in the data. They repeat this process $20$~times, and set $\hat{p}$ to be the average output probabilities over all examples. Given a test example with original output probabilities $p$, they then use the calibrated probabilities $q = \text{softmax}(p / \hat{p})$.

\paragraph{Few-shot fine-tuning}
\label{subsec:few_shot_ft}
Finally, we experiment with few-shot, parameter-efficient fine-tuning for adapting LLMs to a given task's label distribution, thus potentially mitigating label bias. We fine-tune task-specific models for each context prompt using Low-Rank Adapation (LoRA;~\citealp{hu2022lora}), training adapters on 16 held-out training examples for 5 epochs. Importantly, we use the same context $C$ during both fine-tuning and evaluation. Due to computational constraints, we only run LoRA on Llama-2 7B and Mistral 7B, only consider values of $k \in (0,8,16)$, and average
across two sets of demonstrations. See \appref{app_lora} for more details.

%% file: figures/before_mitigation/llama2.tex
\begin{figure*}[t]
    \centering
    \begin{subfigure}[t]{0.31\textwidth}
        \centering
        \includegraphics[width=\textwidth]{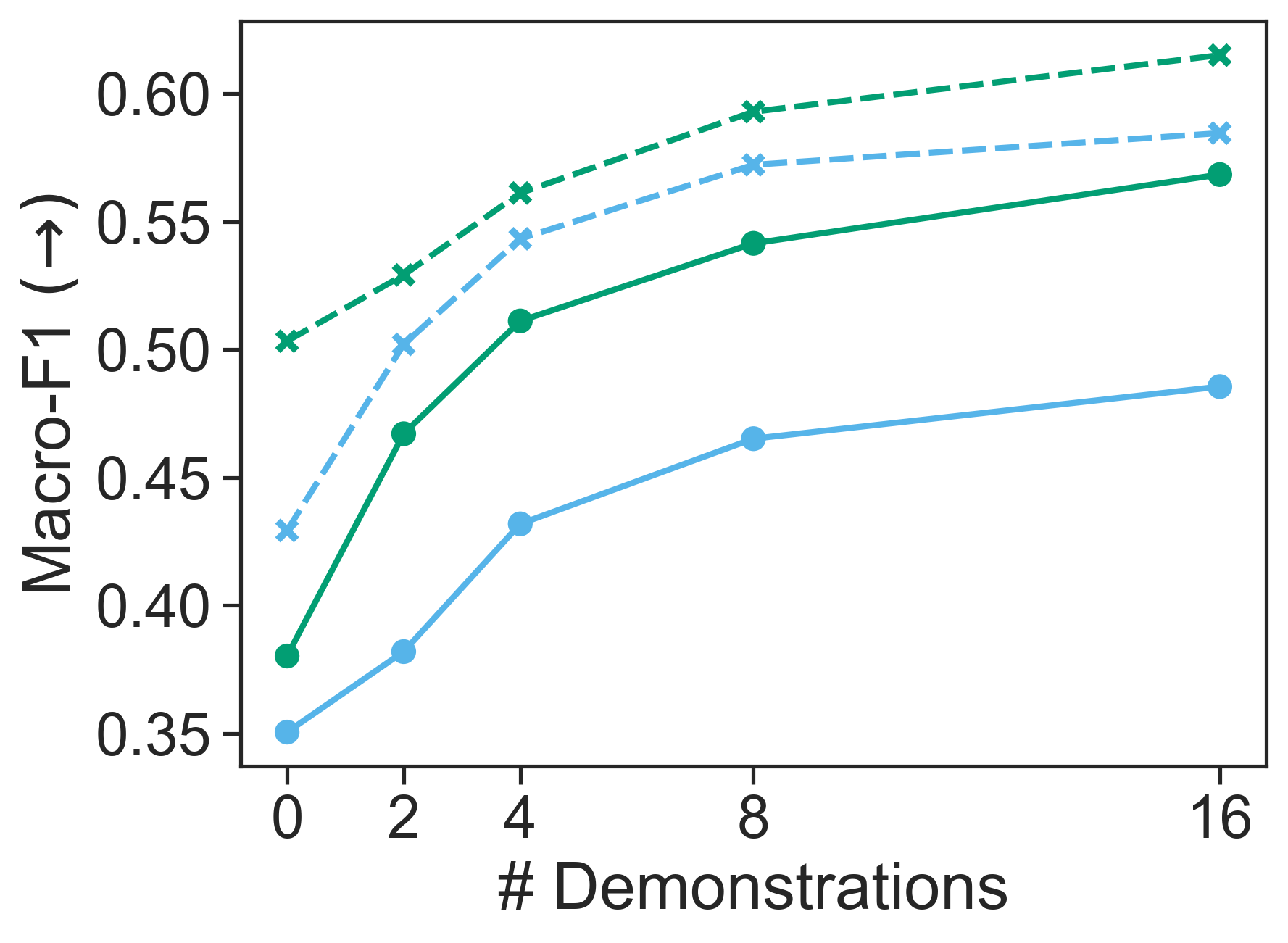}
        \caption{Performance (Macro-F1)}\label{fig:llama2_f1}
    \end{subfigure}%
    ~ 
    \begin{subfigure}[t]{0.31\textwidth}
        \centering
        \includegraphics[width=\textwidth]{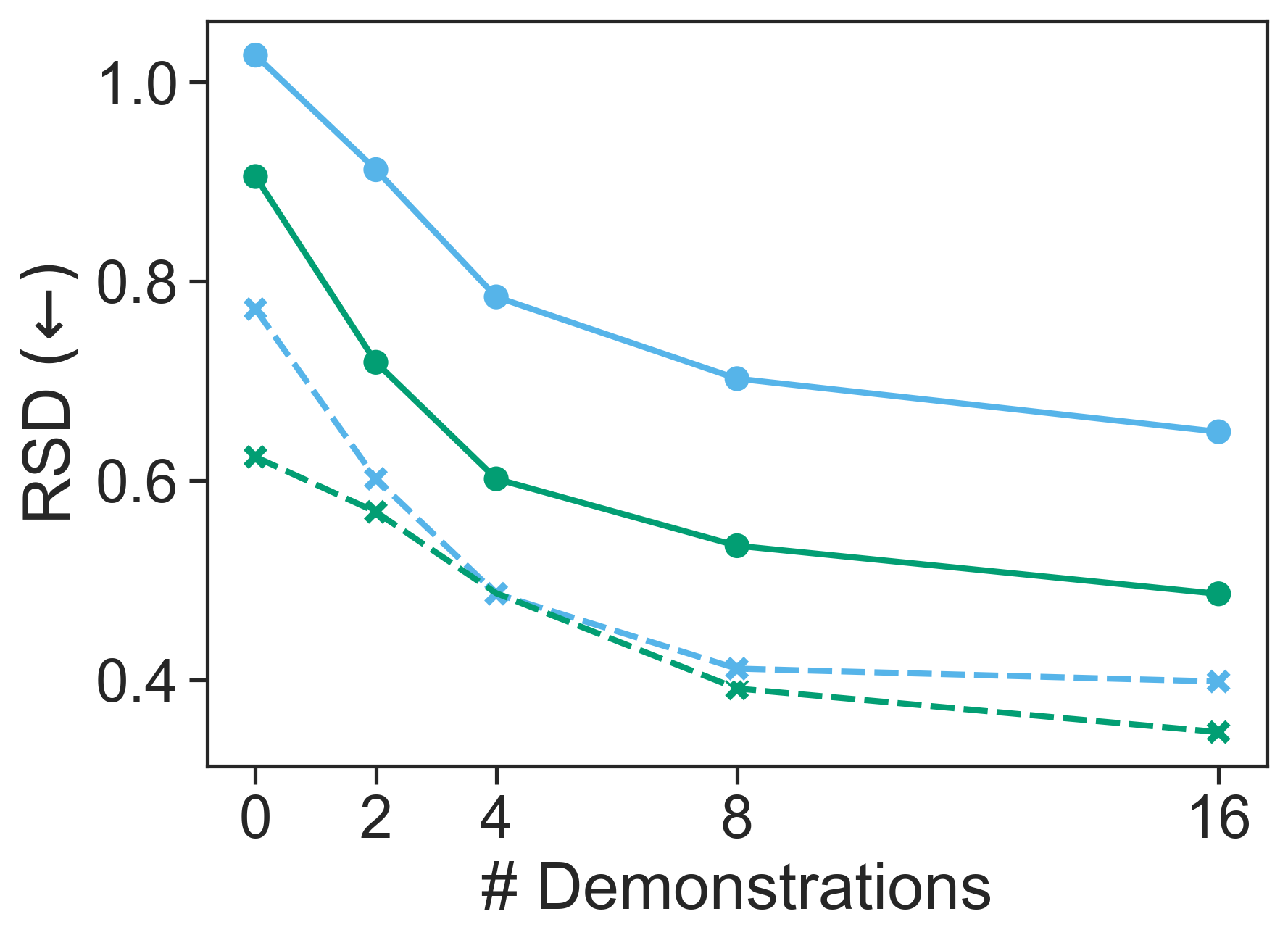}
        \caption{Label bias (\rsd)}\label{fig:llama2_RSD}
    \end{subfigure}
    ~
    \begin{subfigure}[t]{0.31\textwidth}
        \centering
        \includegraphics[width=\textwidth]{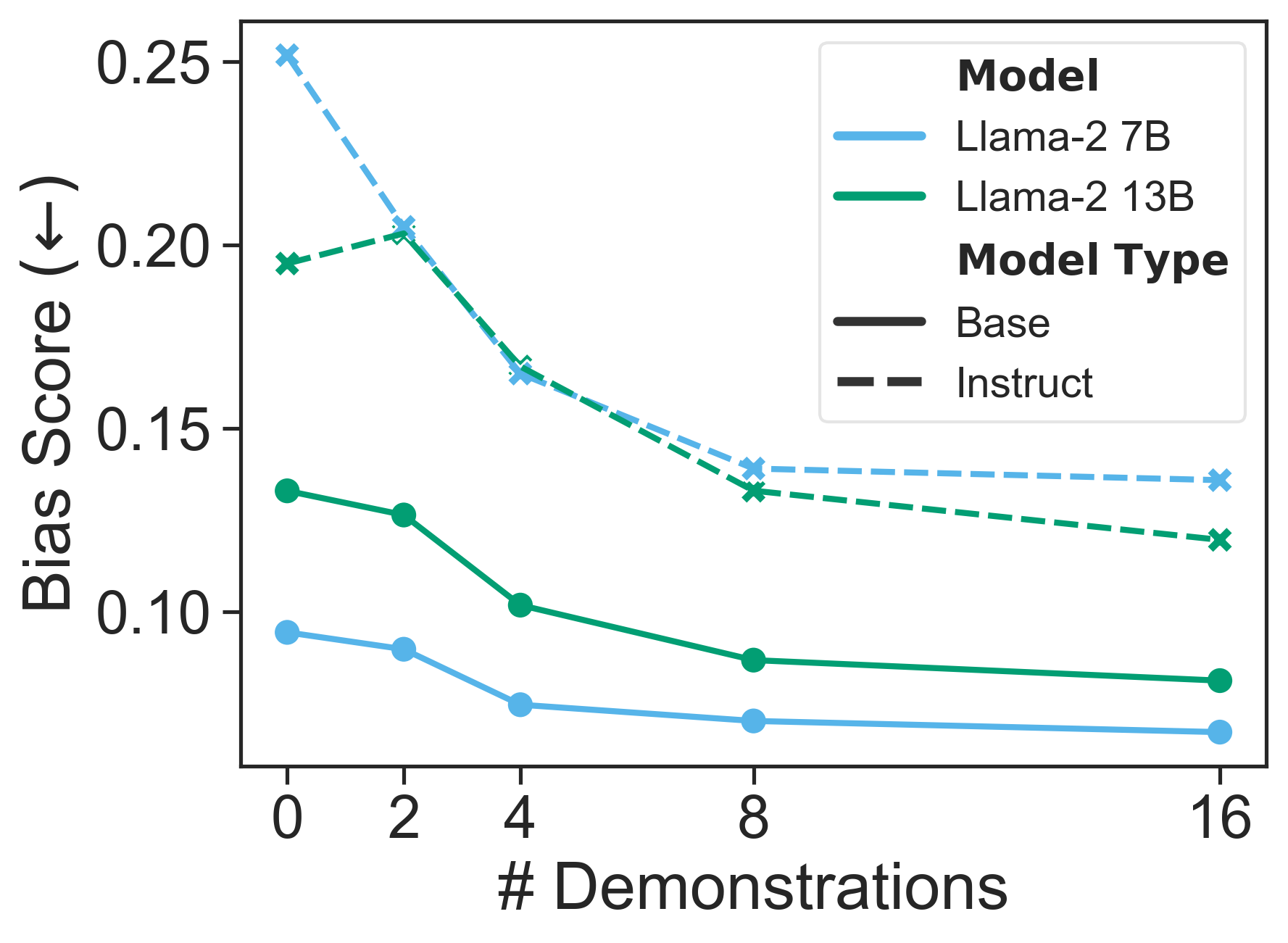}
        \caption{Label bias (\bias)}\label{fig:llama2_BiasScore}
    \end{subfigure}%
    
    \caption{Performance (higher is better) and label bias metrics (lower is better) for Llama-2  pretrained and instruction-tuned models (7B/13B). Both performance and \rsd improve with scale, instruction tuning, and number of demonstrations. 
    In contrast, \bias is substantially worse after instruction tuning and does not improve when scaling models up in most evaluated settings.
    }
\label{fig:llama2_before_mitigation}
\end{figure*}

%% file: text/4.results.tex
\section{Quantifying Label Bias in LLMs}

\subsection{LLMs are Label-Biased}
\label{subsec:llms_are_biased}
We begin by examining the performance and label bias of models with and without instruction-tuning. We report averaged results across all tasks for Llama-2 models in \figref{llama2_before_mitigation}. Results for other models show similar trends (see~\appref{app_before_mitigation_results}).

We first verify that, as expected, model performance~(\figref{llama2_f1}) substantially improves with scale, with instruction tuning and with the number of demonstrations.
We then consider the two bias metrics---\rsd~(\figref{llama2_RSD}) and \bias~(\figref{llama2_BiasScore}). 
We observe that label bias is substantial across most evaluated settings: When prompted with two or no demonstrations, all models obtain high \rsd values of 0.6 or more, with base models obtaining even higher values around 0.9. 
This implies a widespread disparity in model performance across classes in many of the evaluated tasks, and indicates that for most tasks, models primarily succeed on instances of certain classes, while consistently failing on others. Increasing the number of demonstrations to 8 helps reduce the bias, but \rsd remains substantial at around 0.4, and adding further demonstrations results in little to no improvement. 

Similarly, we find \bias improves considerably when using sufficient demonstrations, with models obtaining values as high as 0.25 when using no demonstrations, to around 0.05 for the best model and setting. 
High \bias values indicate the model is uncalibrated, and tends to make overly confident predictions on certain labels regardless of the input. 
Although \bias can be relatively small for some models---indicating their average output distribution is close to uniform---when observed together with high \rsd, it implies that the model subtly but persistently assigns more probability mass to the preferred labels, resulting in substantial bias.

\input{figures/mitigation_methods/llama2}

\subsection{Differences between the Bias Measures}

We further observe that, interestingly, both bias metrics show divergent trends. Although \rsd values, much like model performance, sharply improve after instruction-tuning, the resulting models' \bias is often higher than their base counterparts.
Similarly, while \rsd improves with scaling, the \bias of  smaller models is lower.  

We note that higher performance together with lower \rsd means that the model's performance has improved across most classes. In contrast, higher \bias indicates that its average predicted probabilities grew farther than uniform. Taken together, this implies that the scaled-up and instruction-tuned models are making more confident predictions on some classes, but not on others.  
This could mean more confident correct predictions on the preferred classes, or more confidently wrong predictions on others (or both). 
Altogether, this suggests 
more subtle forms of bias persist after instruction-tuning or scaling up~\cite{tal-etal-2022-fewer}.

Overall, we find the two metrics to be complimentary due to their measurement of different aspects of label bias. We hence use both in further experiments to provide a more comprehensive understanding of label bias in model predictions.

\subsection{Label Bias Persists after Mitigation}
\label{subsec:effects_of_mitigation}

We have seen that LLMs demonstrate extensive label bias across different models, scales and tasks~(\subsecref{llms_are_biased}). We next examine techniques aimed at mitigating such bias, and assess the extent of label bias remaining after their application.
We report our results for Llama-2 models in  \figref{mitigation_methods_llama2}, and observe similar trends for other models~(\appref{app_mitigation_results}).

We first consider the effect of bias mitigation on model performance~(\figref{mitigation_methods_llama2_f1})  using the three methods described in~\subsecref{calibration_methods}: contextual calibration (CC), domain-context calibration (DC), and few-shot fine-tuning with LoRA. 
Compared to standard prompting (\textbf{black} lines), we find that applying CC (\textbf{\textcolor{orange}{orange}}) provides little to no gains. Moreover, it can even undermine model performance, especially for instruction-tuned models, as previously observed by \citet{fei-etal-2023-mitigating}.
In contrast, DC (\textbf{\textcolor{Thistle}{purple}}) can provide substantial performance gains, especially when using few or no in-context demonstrations, where baseline performance is relatively low. However, when calibrating instruction-tuned models prompted with a higher number of demonstrations, we find that DC mostly fails to improve performance.
Finally, LoRA considerably improves performance in all cases~(\textbf{\textcolor{Green}{green}} in \figref{mitigation_methods_llama2}, upper row), vastly outperforming both CC and DC.

We next turn to measure label bias~(\figref{mitigation_methods_llama2_RSD} and~\ref{fig:mitigation_methods_llama2_BiasScore}).
Notably, here we observe that for the two calibration methods, changes in both \rsd and \bias are correlated with changes in performance.
We find that CC substantially worsens label bias in instruction-tuned models, and can also increase bias for base models. Conversely, while DC alleviates bias in many of the evaluated settings, it is largely unsuccessful in mitigating it when prompting instruction-tuned models with $8$ or more demonstrations. LoRA proves effective for improving \rsd in all settings, but \rsd values still remain relatively high. In contrast, \bias noticeably increases after LoRA fine-tuning, indicating that more subtle bias persists.

Overall, our results indicate that existing bias calibration approaches are insufficient for diminishing label bias in essential cases, particularly for instruction-tuned models. Further, while LoRA fine-tuning is effective in both improving performance and mitigating certain aspects of bias (though not others), 
it is also considerably more computationally expensive than calibration.

%% file: figures/mitigation_methods/llama2.tex
\begin{figure*}[t]
    \centering
    \captionsetup[subfigure]
    {font=small,labelfont=small} 
    
    \begin{subfigure}[b]{0.03\textwidth}
        \centering
        \mbox{}
        \vfill
        \rotatebox[origin=c]{90}{\small 7B}
        \bigskip
        \vfill
        \vspace{1.1cm}
        \mbox{}
    \end{subfigure}%
    \begin{subfigure}[b]{0.31\textwidth}
        \centering
        
        \includegraphics[width=\textwidth]{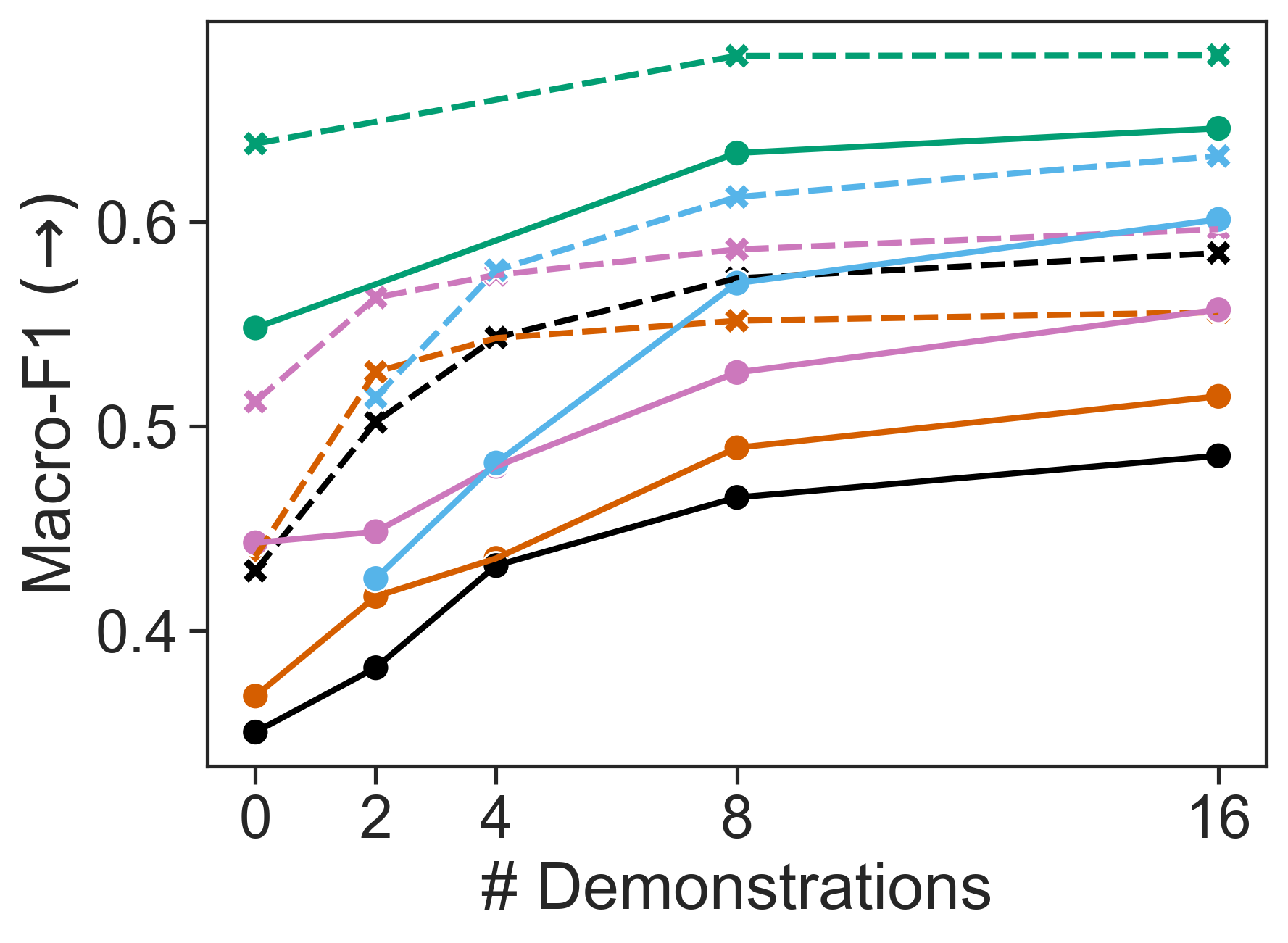}
    \end{subfigure}%
    ~ 
    \begin{subfigure}[b]{0.31\textwidth}
        \centering
        
        \includegraphics[width=\textwidth]{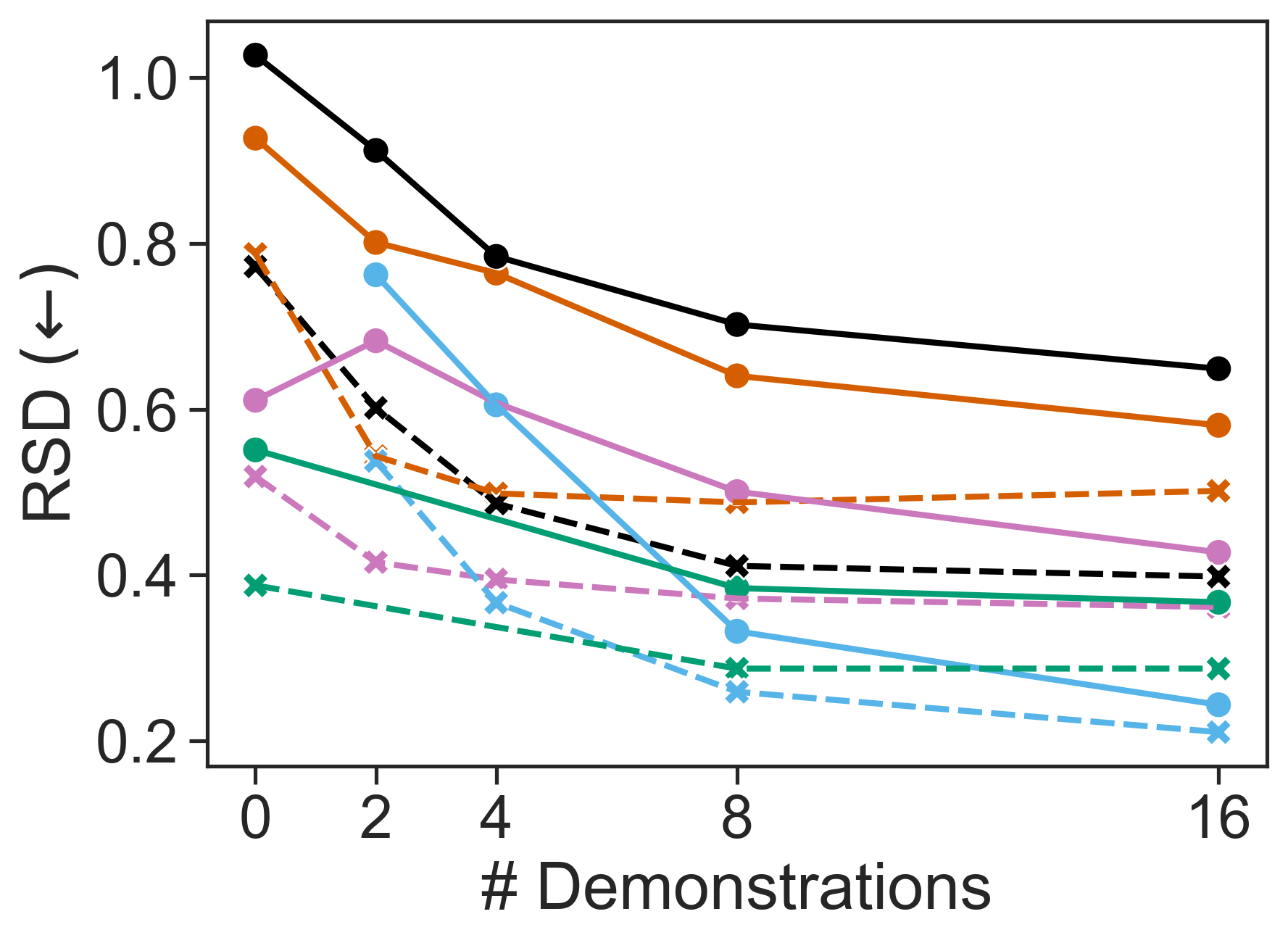}
    \end{subfigure}
    ~
    \begin{subfigure}[b]{0.31\textwidth}
        \centering
                
        \includegraphics[width=\textwidth]{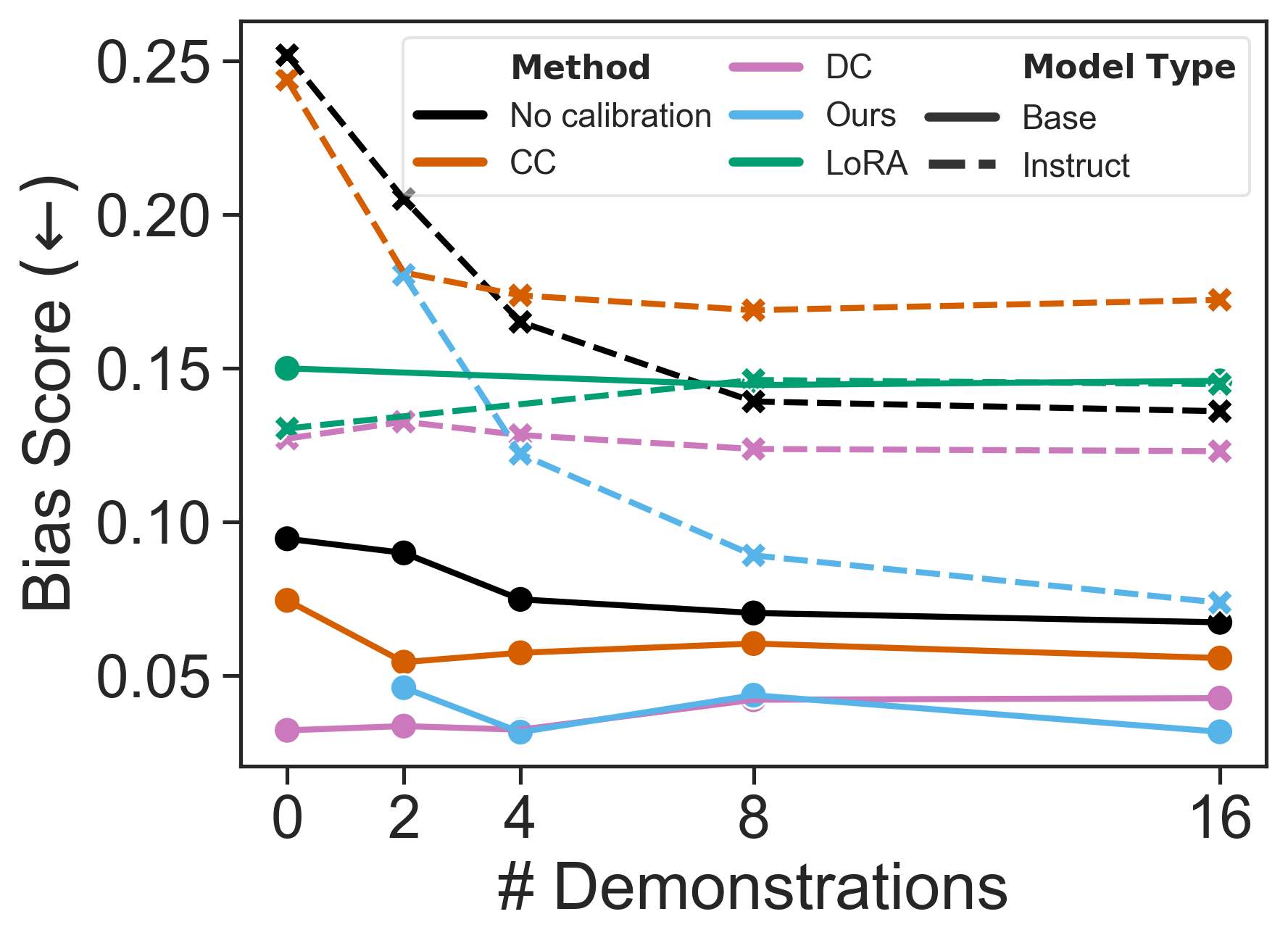}
    \end{subfigure}%

    
    \begin{subfigure}[b]{0.03\textwidth}
        \centering
        \mbox{}
        \vfill
        \rotatebox[origin=c]{90}{\small 13B}
        \bigskip
        \vfill
        \vspace{1.7cm}
        \mbox{}
    \end{subfigure}%
    \begin{subfigure}[b]{0.31\textwidth}
        \centering
        
        \includegraphics[width=\textwidth]{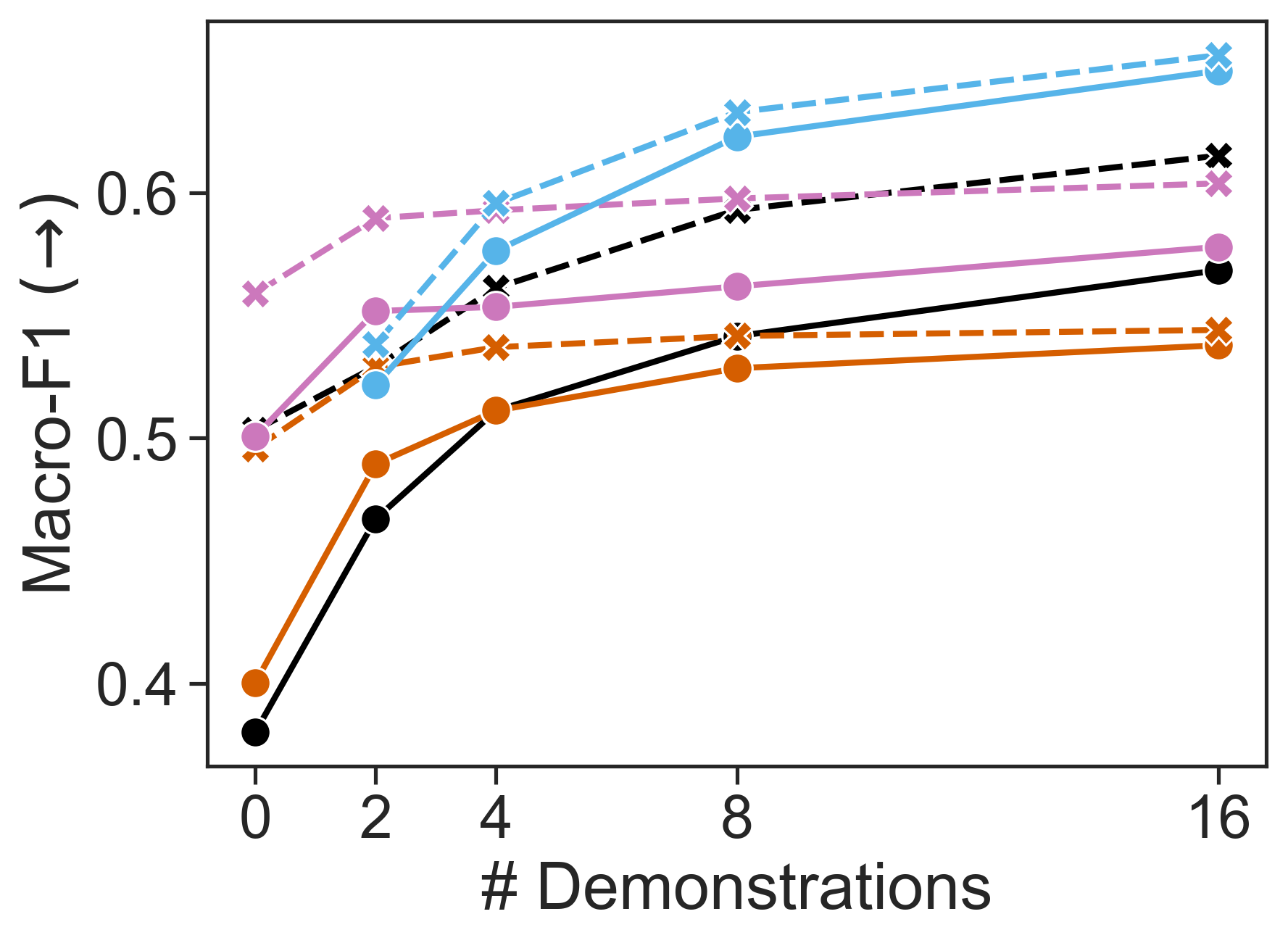}
        
        \subcaption{\hspace{0.5cm}\textsl{Macro-F1}}
        \label{fig:mitigation_methods_llama2_f1}
        
    \end{subfigure}%
    ~ 
    \begin{subfigure}[b]{0.31\textwidth}
        \centering
        
        \includegraphics[width=\textwidth]{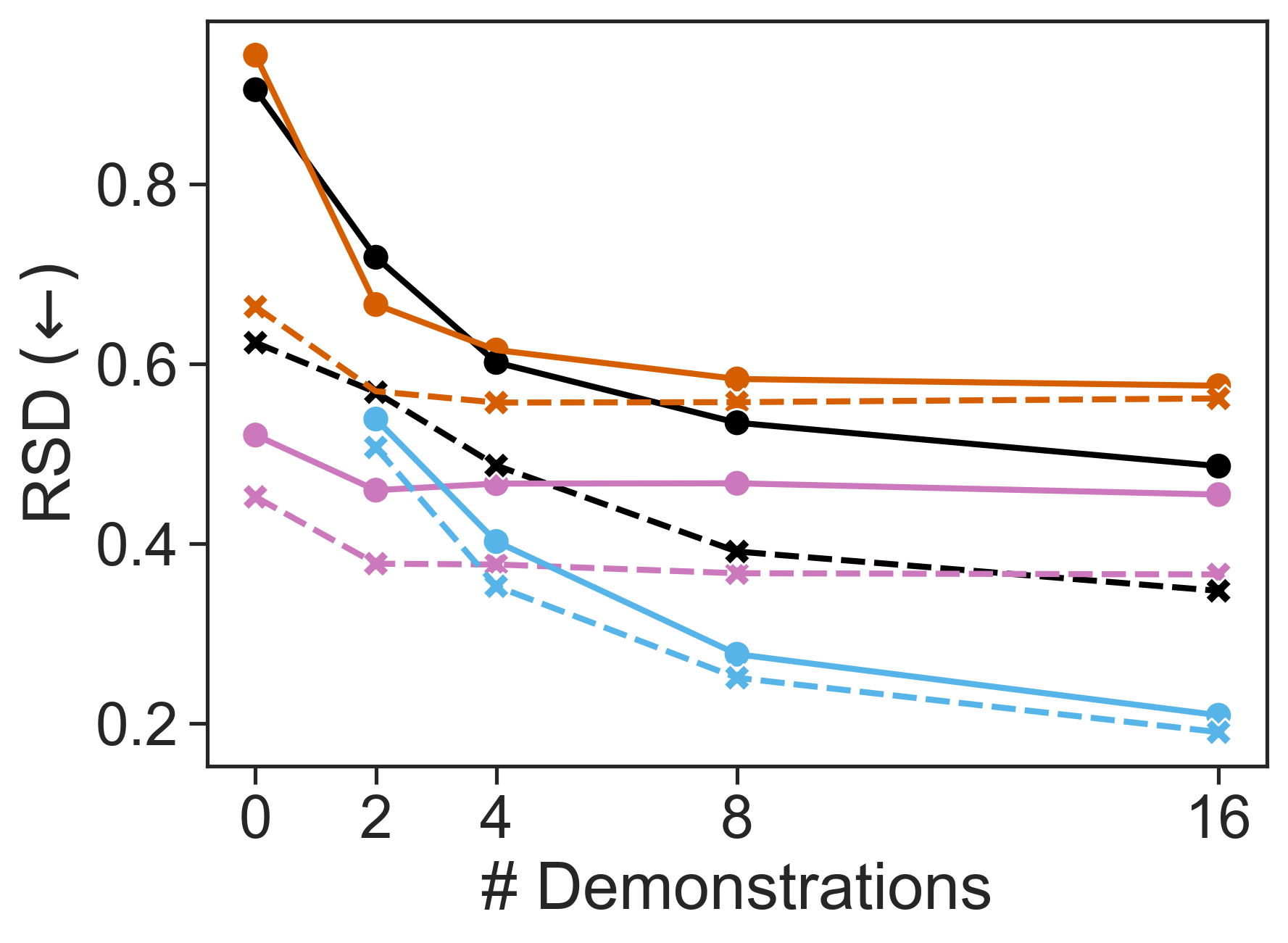}

        \subcaption{\hspace{0.5cm} \textsl{RSD}}
        \label{fig:mitigation_methods_llama2_RSD}

    \end{subfigure}
    ~
    \begin{subfigure}[b]{0.31\textwidth}
        \centering
        
        \includegraphics[width=\textwidth]{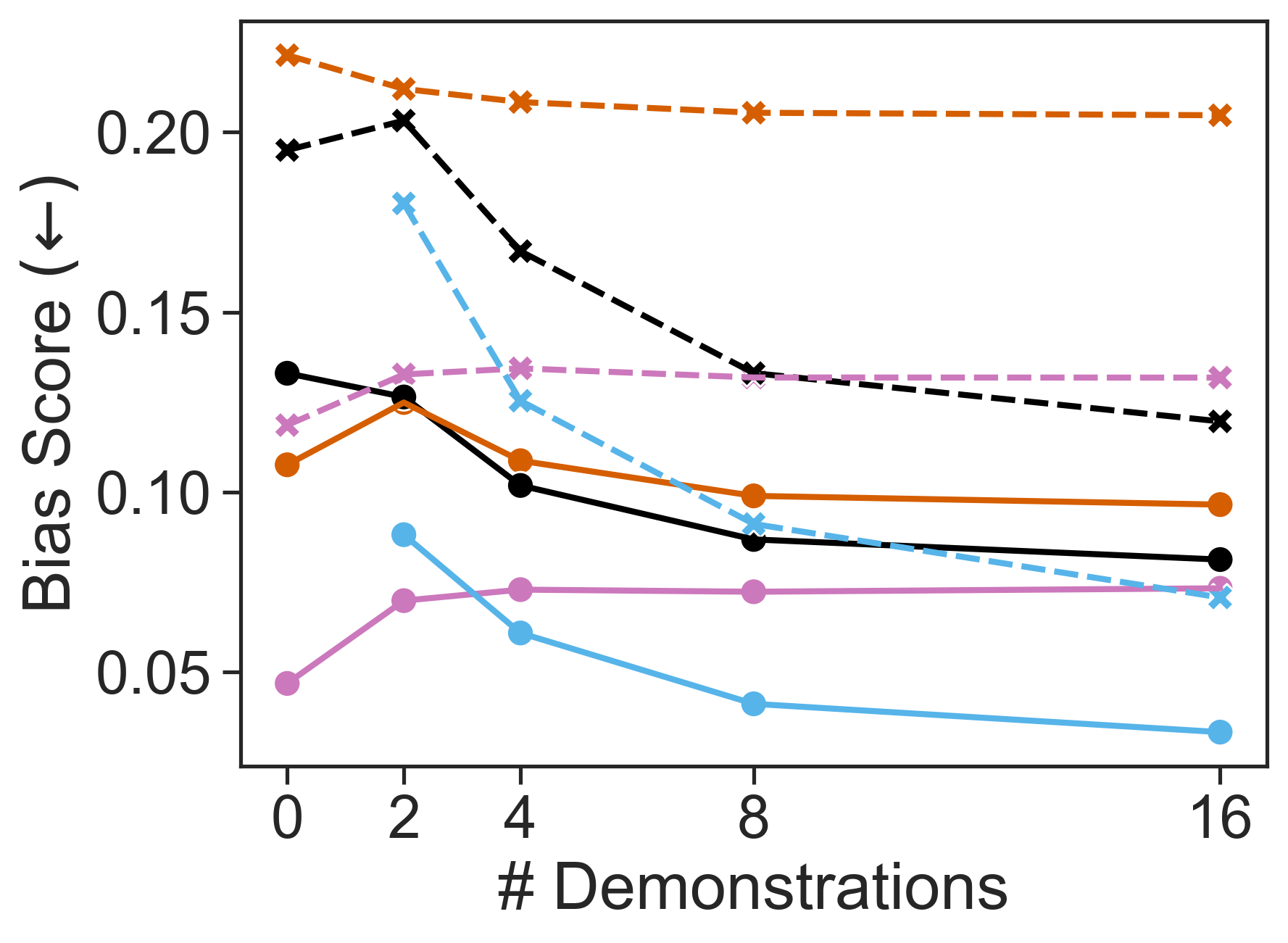}

        \subcaption{\hspace{0.5cm} \textsl{Bias Score}}
        \label{fig:mitigation_methods_llama2_BiasScore}

    \end{subfigure}%

    \caption{The effect of label bias mitigation methods on performance and bias for Llama-2 models. CC improves neither performance nor bias; DC and LoRA fine-tuning improve both; our \textsl{Leave-One-Out Calibration} (LOOC) method leads to the best  performance among the calibration methods, and the overall lowest bias for $k\in\{8,16\}$.} 
    
\label{fig:mitigation_methods_llama2}
\end{figure*}

%% file: text/5.calibration.tex
\section{Mitigating Label Bias by Calibrating on Demonstrations} 
\label{sec:looc}
Motivated by the challenges of existing calibration approaches on instruction-tuned models (\subsecref{effects_of_mitigation}), we aim to develop an effective calibration method for such scenarios. 
We hypothesize a possible cause for such difficulties is that the inputs used for calibration in CC and DC are very distinct from the more curated, high-quality inputs models observe during instruction-tuning \cite{touvron2023llama}.\footnote{Specifically, nonsensical task inputs made up of random words as in DC, or placeholder-like strings as in CC, are less likely to be observed during instruction tuning.}

Seeking to use more naturally-occurring inputs, and to avoid any reliance on additional held-out examples, we propose to calibrate models using the in-context examples readily available in few-shot prompts. 
We therefore need to obtain the model's output probabilities on these inputs to estimate its bias.
However, as these examples appear alongside their labels in the context provided to the model, it could simply copy the correct answer from the prompt, leading to unreliable bias estimates.
We introduce a simple method to alleviate this concern.

\paragraph{Leave-One-Out Calibration (LOOC)} Our goal is to estimate the model's average output probabilities $\hat{p}$ at test-time by using the $k$ demonstrations $[(x^1, y^1), \ldots, (x^k, y^k)]$ provided in the context $C$, and then use it for calibration. 
Drawing from leave-one-out cross-validation, when evaluating the model on the $i$-th demonstration's input $x^i$, we prompt it with an edited context $C_{-i}$ comprised of the original context $C$ after removing the current demonstration $(x^i, y^i)$, resulting in $k-1$ demonstrations.\footnote{We leave all other demonstrations in their original order.}
We thus obtain $k$ output probabilities: 
\begin{equation*}
    p^i(y) = p(y \mid x^i, C_{-i})
\end{equation*}
To reliably estimate $\hat{p}$, we further need to account for the demonstrations' labels $y^i$: for imbalanced choices of demonstrations (e.g., class imbalance), using the average of $p^i$'s could lead to an underestimation of the probability assigned to infrequent labels.
We therefore compute the average output probabilities $\hat{p}$ by taking into account the labels $y^i$, as we do for computing \bias (\subsecref{metrics}): We first average $p^i$'s associated with the same label $\ell$, $\mathcal{D}_\ell = \{ p^i \mid y^i = \ell \}$, and then set $\hat{p}$ as the mean of these intra-label averages:
\begin{equation*}
    \hat{p}(y) = \frac{1}{|Y|} \sum_{\ell \in Y}  \left( \frac{1}{\left| \mathcal{D}_\ell \right|} \sum_{p^i \in 
    \mathcal{D}_\ell} p^i(y) \right)
\end{equation*}
 Finally, we use $\hat{p}$ to compute calibration parameters and score new examples using the same methodology as \citealp{zhao2021calibrate} (\subsecref{mitigation_methods}).
We refer to our method as Leave-One-Out Calibration (LOOC).

\paragraph{Results}
We use LOOC to calibrate models in the same setup of~\subsecref{effects_of_mitigation}. We report our results for Llama-2 models in \figref{mitigation_methods_llama2} (\textbf{\textcolor{Cyan}{cyan}} lines), finding similar trends in other models (\appref{app_mitigation_results}). 
Comparing our method to other calibration approaches, we find LOOC surpasses CC and DC by a wide margin in both performance and bias metrics for prompts with $k=8,16$ demonstrations. Importantly, using LOOC to calibrate instruction-tuned models in this setting dramatically improves upon the uncalibrated model, whereas other calibration methods fail to achieve meaningful gains (\subsecref{effects_of_mitigation}).
Further, LOOC nearly closes the gap with LoRA-level performance while improving upon it in both bias metrics, yet uses substantially less compute.

As LOOC relies on the in-context demonstrations for bias estimation, $k$ needs to be sufficiently large for calibration to succeed. Surprisingly, we find that with as few as $k=4$ demonstrations, our method is often comparable to the next best calibration method on all metrics.
Finally, we note that while our method can substantially reduce label bias compared to other approaches, the remaining \rsd is still considerable, indicating that model performance is still biased on some tasks.

%% file: text/6.analysis.tex
\section{Analysis}
We study the effect of different factors on the extent of label bias in model predictions: the semantic meaning of the task labels~(\subsecref{sem_eq_labels}), the level of label imbalance in the demonstrations~(\subsecref{level_of_imbalance}), and the choice of demonstrations~(\subsecref{bias_sensitivity}).

\subsection{Semantically Equivalent Labels}
\label{subsec:sem_eq_labels}

The output space for classification tasks often consists of labels with strong semantic meaning (e.g., ``Positive'' vs.~``Negative''). Recent work has indicated that, when faced with such labels, models are affected by semantic priors from their pretraining or instruction-tuning~\cite{wei2023larger, min-etal-2022-rethinking} that could affect label bias~\cite{fei-etal-2023-mitigating}.

We examine whether models exhibit lower label bias when the task's labels are semantically equivalent and interchangeable.
We extract all multi-choice QA tasks---with label spaces such as ``A/B/C/D'' or ``1/2/3''---and all sentence completion tasks, where models choose a logical continuation for an input text between two options, usually labeled A and B.
This results in $18$ tasks with semantically equivalent labels.

We compare label bias on this subset of tasks and the entire evaluation suite for Llama-2 models in \figref{sem_eq_labels_llama2}, with results for other models largely following similar trends~(\appref{app_sem_eq_labels}). 
We find that, in most cases, models demonstrate lower label bias on tasks with semantically equivalent labels. This is especially evident in settings with few or no demonstrations, where models are typically strongly biased~(\subsecref{llms_are_biased}). Still, \rsd levels for such tasks remain relatively high across all evaluated settings. Further, we observe that instruction-tuned models prompted with 8 or more demonstrations are often \emph{more} biased on this subset of tasks. In summary, although using semantically equivalent labels may potentially mitigate bias in scenarios with limited demonstrations, LLMs still exhibit substantial label bias when faced with such labels.

\subsection{Imbalanced In-context Demonstrations}
\label{subsec:level_of_imbalance}

Label imbalance in the in-context demonstration set was previously shown to amplify label bias~\cite{zhao2021calibrate} as well as decrease model performance~\cite{min-etal-2022-noisy}, but such results were derived on a restricted set of tasks. 
We use our evaluation suite to investigate the observed label bias and performance of models when varying the level of imbalance in the demonstrations. To establish a consistent definition of label imbalance across different tasks, we use the subset of binary classification tasks ($N=197$) with $k=8$ demonstrations. 
Given a task with labels $L=\{\ell_A, \ell_B\}$ and a context $C$, we define $p_{\uparrow}$ as the proportion of the most frequent label in the demonstrations of $C$, such that $p_{\uparrow}$ attains values in $\{0.5, 0.625, 0.75, 0.875, 1.0\}$. Specifically, $p_{\uparrow}=0.5$ means the labels are perfectly balanced, and $p_{\uparrow}=1.0$ means the demonstrations only include examples for one of the labels.

\input{figures/sem_eq_labels/llama2}

\input{figures/imbalanced_labels/fig_v2}

For every task, we prompt Llama-2 (7B/13B) and Mistral (7B) models using 10 different sets of demonstrations, with 2 sets for each value of $p_{\uparrow}$: one where $\ell_A$ is the most frequent label in $C$, and another where $\ell_B$ is the most frequent, 
as well as two different balanced sets ($p_{\uparrow}=0.5$).\footnote{To build each set, we randomly select and permutate 8 demonstrations from a pool of 16 held-out examples, while controlling for the selected number of examples per label.} 
We group measurements taken across different tasks and demonstration sets by their level of label imbalance $p_{\uparrow}$, and inspect the average results per level.

We report our results in~\figref{imbalanced_labels}. Examining the two bias metrics, \rsd~(\figref{imbalanced_labels_RSD}) and \bias~(\figref{imbalanced_labels_BiasScore}), we observe that both pretrained and instruction-tuned models are resistant to label imbalance:
Increased imbalance does not result in notable gains in bias, unless the imbalance is very extreme---specifically, when the demonstrations include only a single or no demonstrations for one of the labels ($p_{\uparrow}>0.75$).
Interestingly, model performance follows the same trends~(\figref{imbalanced_labels_F1}).
Overall, our results indicate that for most tasks, the impact of label imbalance in the demonstrations set is minimal, except for cases of severe imbalance.

\subsection{Choice of Demonstrations}
\label{subsec:bias_sensitivity}

The performance of LLMs in in-context learning was shown to be sensitive to the exact choice of demonstrations used to prompt the model~\cite{liu-etal-2022-makes, chang-jia-2023-data}. We examine whether such choices also impact the extent of \emph{label bias} in model predictions.
We assess the performance and bias of Llama-2 (7B/13B) and Mistral (7B) models across 5 different sets of $k=8$ demonstrations for each task in our evaluation suite. In addition to reporting the mean and standard deviation of each metric, we use several oracle methods to aggregate and choose a specific demonstration set per task when computing the overall cross-task performance and bias metrics. Specifically, we select the demonstration sets that attain the following, per task: \emph{best performance}; \emph{worst performance}; \emph{median performance};  \emph{least bias}; and \emph{most bias}.

\input{tables/llama2-7b-base_icl_set}

We report our results for Llama-2 7B base in \tabref{llama2-7b-base_icl_set_choice}, with other models showing similar trends~(\appref{app_bias_sensitivity}).
We find that label bias, similarly to model performance, is highly sensitive to the choice of demonstrations, as indicated by the high variance across sets. Interestingly, the set of demonstrations that attains the worst performance also leads to strong bias, and vice-versa. In fact, we find that performance and bias are anti-correlated, with strong Pearson correlation for \rsd ($r=-0.74$) and moderate for \bias ($r=-0.30$), 
indicating that when LLMs underperform in classification, it is often due to prompts that exacerbate bias. We leave further research into demonstrations that lead to biased and unbiased predictions to future work.

%% file: figures/sem_eq_labels/llama2.tex
\begin{figure}[t]
    \centering
    \captionsetup[subfigure]
    {font=small,labelfont=small} 
    
    \begin{subfigure}[t]{0.68\columnwidth}
        \centering
        
        \includegraphics[width=\columnwidth]{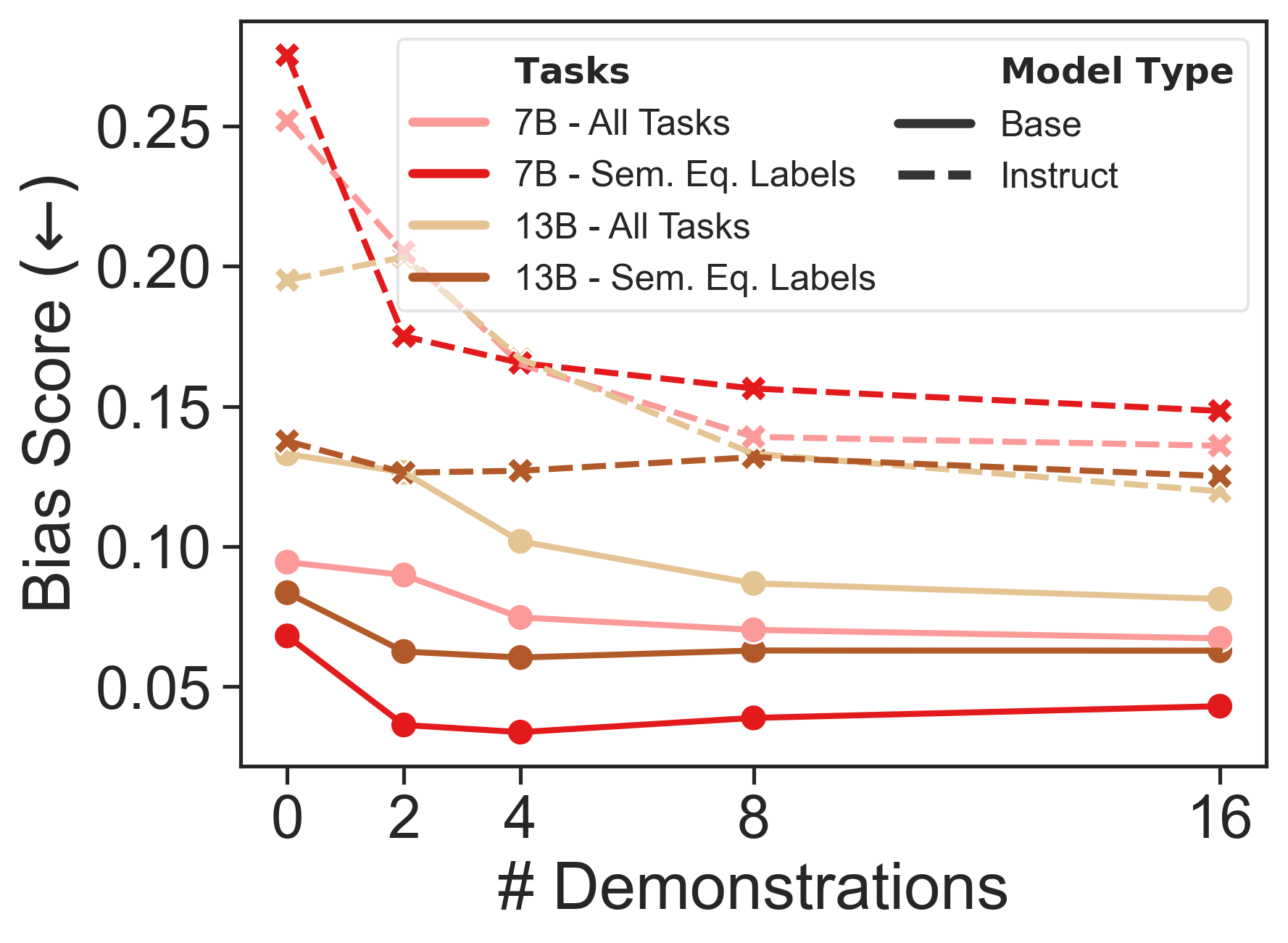}

        \subcaption{\bias}
    \end{subfigure}%
 
    \begin{subfigure}[t]{0.68\columnwidth}
        \centering
        
        \includegraphics[width=\columnwidth]{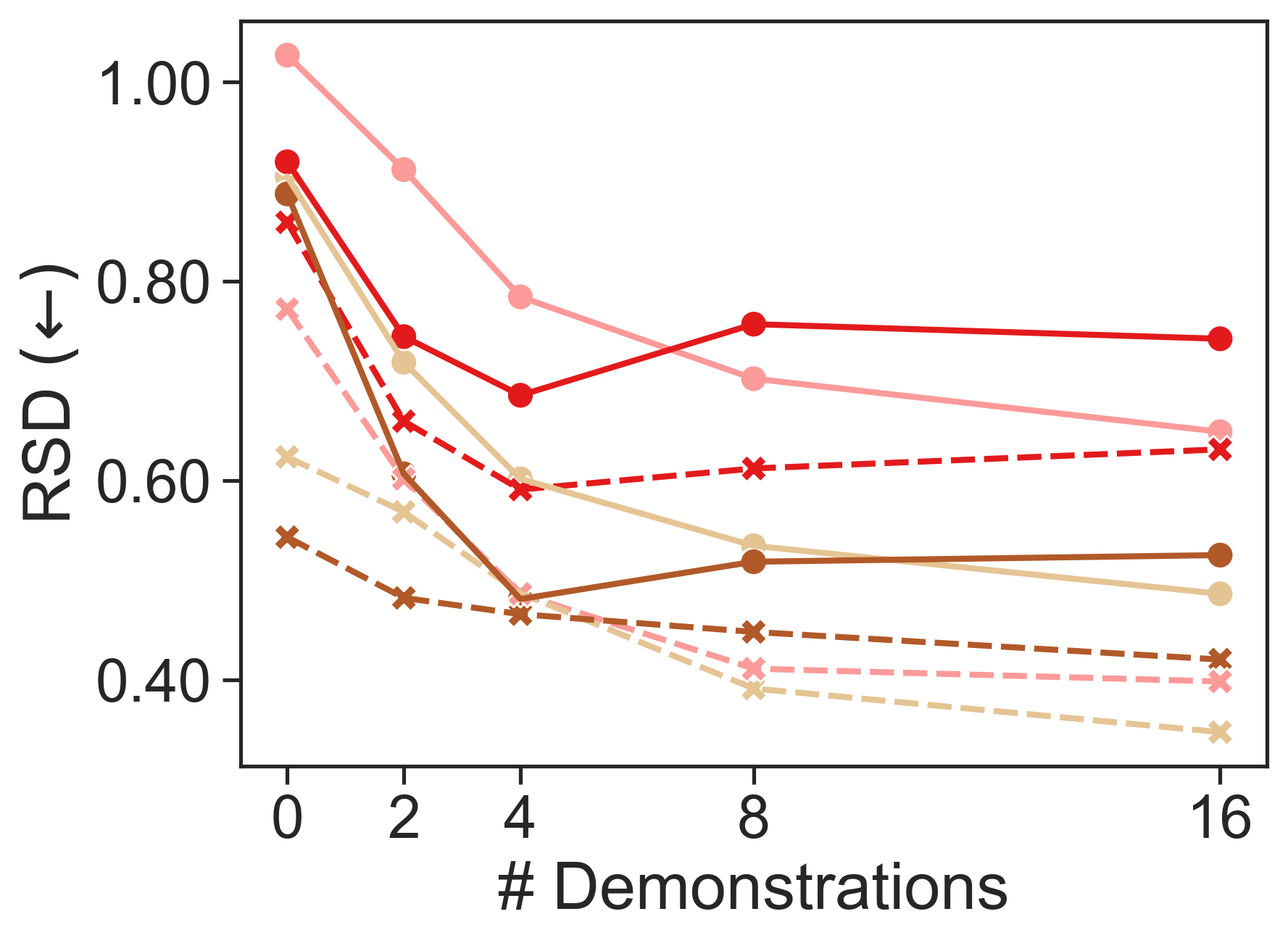}
        
        \subcaption{\rsd}
    \end{subfigure}%
    ~
    \caption{Label bias metrics for Llama-2 models (7B/13B), when evaluated on all tasks in our evaluation suite (\emph{All}) vs.~a subset of tasks with semantically equivalent labels (\emph{Sem.~Eq.~Labels}). LLMs exhibit label bias even on tasks with semantically equivalent labels, such as multi-choice question answering.
    }
    
\label{fig:sem_eq_labels_llama2}
\end{figure}

%% file: figures/imbalanced_labels/fig_v2.tex
\begin{figure*}[t]
    \centering
    \captionsetup[subfigure]
    {font=small,labelfont=small} 
    
    \begin{subfigure}[b]{0.29\textwidth}
        \centering

        \includegraphics[width=\textwidth]{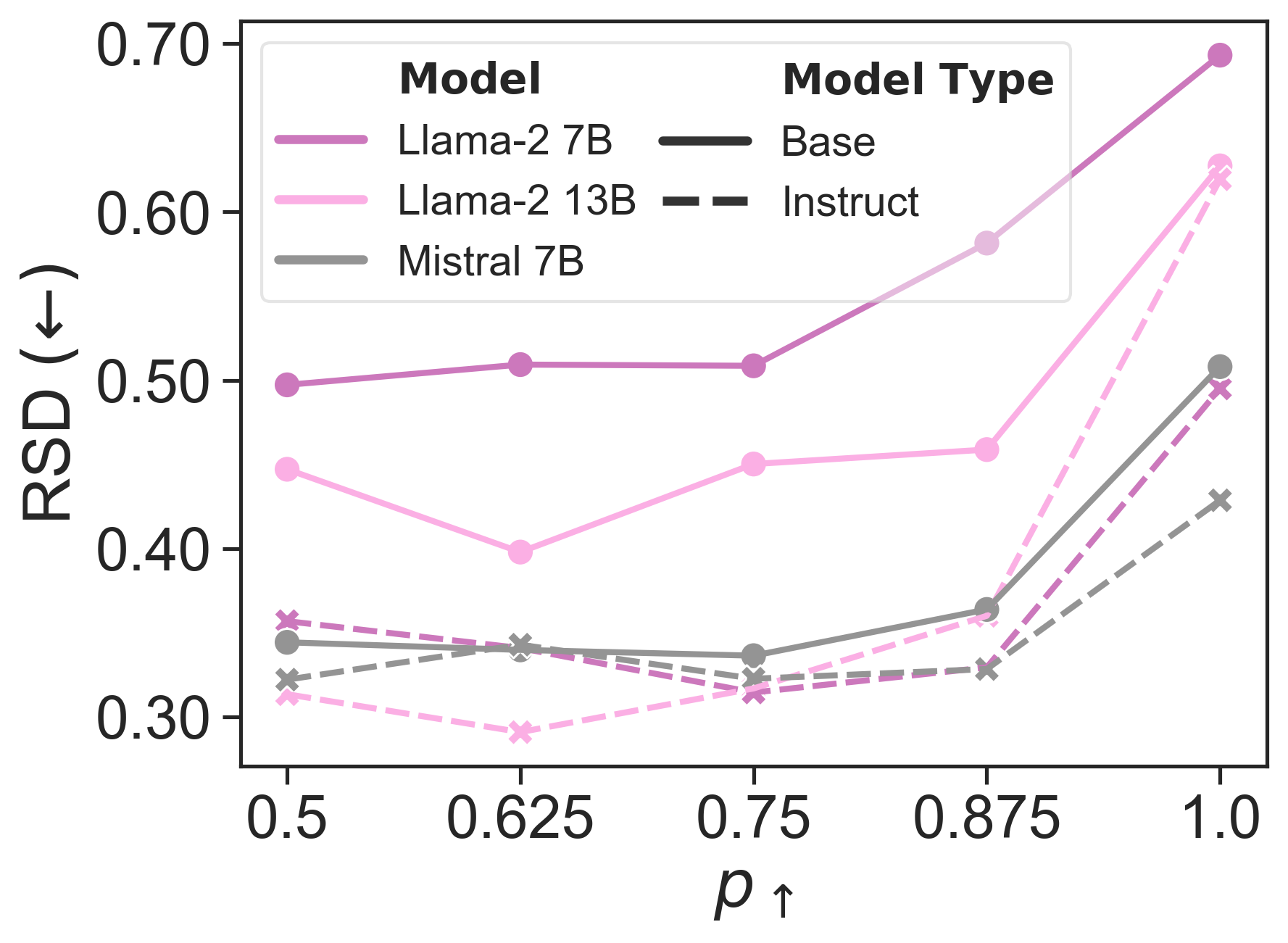}%
        
        \subcaption{\hspace{0.5cm}\rsd}\label{fig:imbalanced_labels_RSD}%
        
        
        
    \end{subfigure}%
    ~ 
    \begin{subfigure}[b]{0.29\textwidth}
        \centering

        \includegraphics[width=\textwidth]{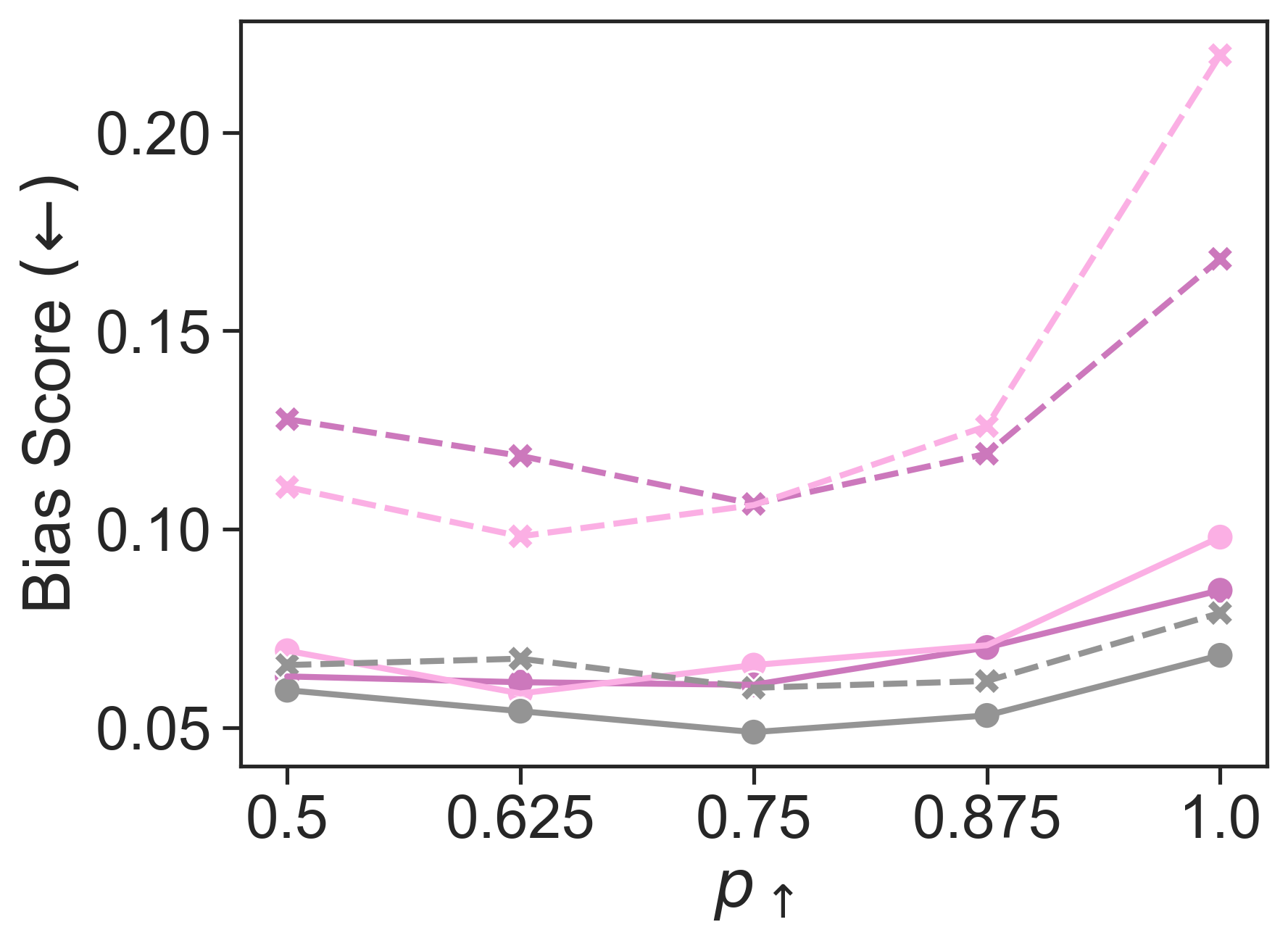}%
        \subcaption{\hspace{0.5cm}\bias}\label{fig:imbalanced_labels_BiasScore}%
        


    \end{subfigure}
    ~
    \begin{subfigure}[b]{0.29\textwidth}
        \centering

        \includegraphics[width=\textwidth]{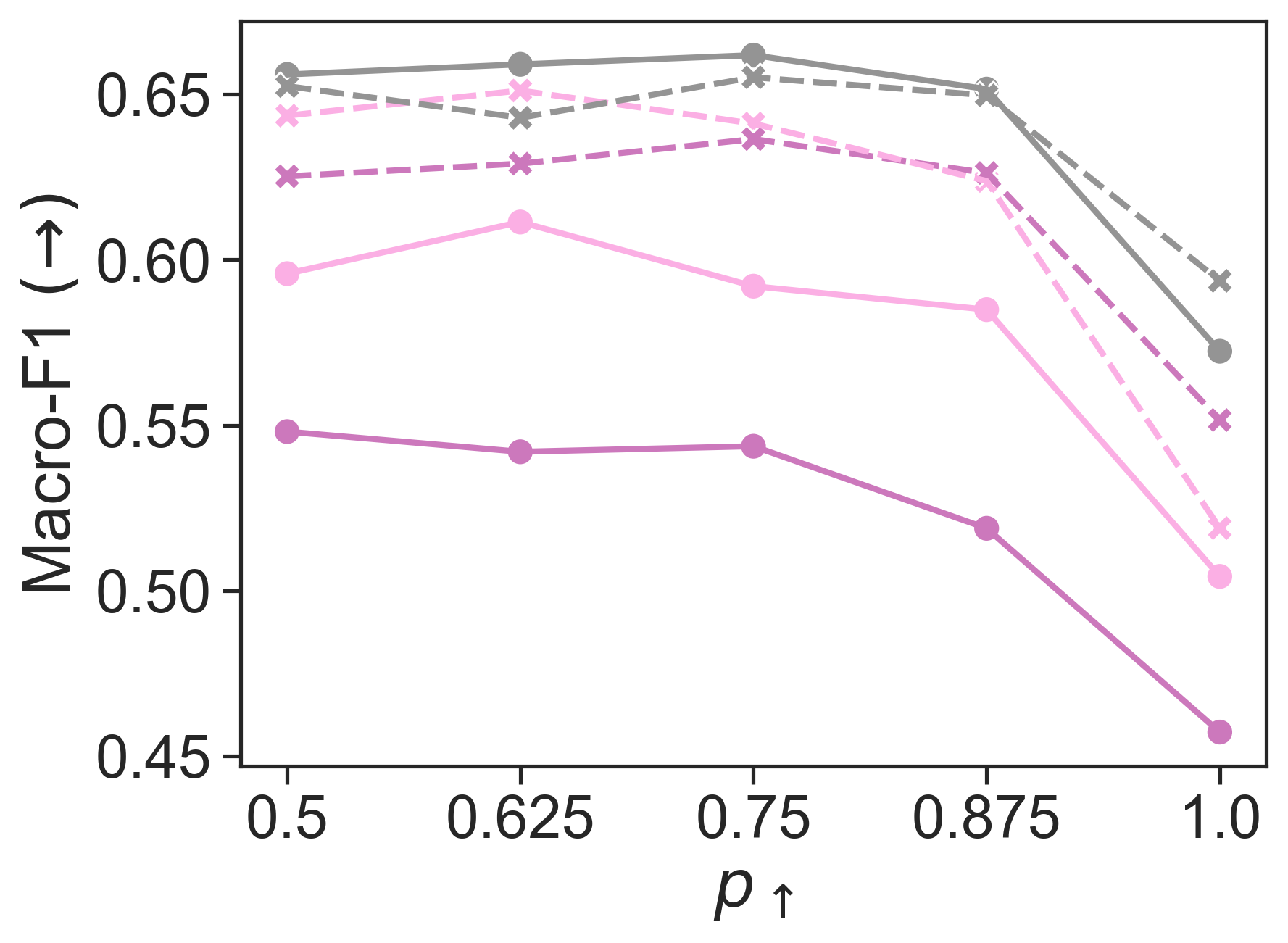}%
        
        \subcaption{\hspace{0.5cm}\textsl{Macro-F1}}\label{fig:imbalanced_labels_F1}%
        


    \end{subfigure}%

\caption{Label bias and performance metrics for Llama-2 (7B/13B) and Mistral (7B) models, when aggregated by the level of imbalance in the demonstrations set used for prompting the model, measured by the proportions of its most frequent label ($p_\uparrow$). For most tasks, label imbalance has only minor impact on both bias and performance, unless the imbalance is extreme. Instruction-tuned models are less sensitive to imbalance. 
}
\label{fig:imbalanced_labels}
\end{figure*}

%% file: tables/llama2-7b-base_icl_set.tex
\begin{table}[tb]
\centering
\resizebox{\columnwidth}{!}{%
\begin{tabular}{@{}cccc@{}}
\toprule
\textbf{Choice of Demonstrations} & \textbf{F1} ($\uparrow$)           & \textbf{\rsd} ($\downarrow$)         & \textbf{\bias} ($\downarrow$)         \\ \midrule
Mean (SD)                     & 0.47 ($\pm$ 0.088) & 0.69 ($\pm$ 0.316) & 0.077 ($\pm$ 0.039) \\ \midrule
Best Performance              & 0.565        & 0.384        & 0.068         \\
Median Performance              & 0.478        & 0.656        & 0.079         \\
Worst Performance             & 0.355        & 1.083        & 0.071         \\ \midrule
Least Bias -- \textsl{by} \rsd             & 0.553        & 0.35         & 0.066         \\
Least Bias -- \textsl{by} \bias            & 0.457        & 0.755        & 0.024         \\
Most Bias -- \textsl{by} \rsd              & 0.358        & 1.102        & 0.069         \\
Most Bias -- \textsl{by} \bias             & 0.436        & 0.781        & 0.119         \\ \bottomrule
\end{tabular}

}

\caption{
Results of Llama-2 7B base model when prompted with 5 different sets of demonstration on our evaluation suite. We employ oracles to aggregate and compute cross-task results by choosing a specific set of demonstrations for each task. Label bias is highly sensitive to the choice of in-context examples.
}
\label{tab:llama2-7b-base_icl_set_choice}
\end{table}

%% file: text/_1.related_work.tex
\section{Related Work}

\paragraph{Biases in LLM predictions} Recent work has revealed various biases in the predictions of LLMs. \citet{wang2023large} showed that models exhibit positional bias when presented with several texts for evaluation and ranking. \citet{pezeshkpour2023large} and \citet{zheng2023large} exposed a similar bias in multi-choice QA. \citet{si-etal-2023-measuring} studied inductive biases in in-context learning.
Complimentary to these works, we study label bias and seek to improve its evaluation and mitigation.

\paragraph{Calibrating Label Bias in LLMs}
Recent work introduced calibration methods to mitigate label bias in LLMs~\cite{zhao2021calibrate,fei-etal-2023-mitigating}. \citet{han2023prototypical} proposed to fit a Gaussian mixture to the model's output probabilities and use it for calibration, but their approach requires hundreds of labeled examples. Concurrently to our work,  \citet{jiang2023generative} proposed to generate inputs for calibration by conditioning models on the context prompt, and \citet{zhou2023batch} calibrate models using model output probabilities on the entire test set. While the motivation for both methods is similar to ours, our approach does not require access to the test set, or any compute to obtain inputs for calibration.
Importantly, unlike previous work on bias calibration, our main focus is the evaluation of label bias in LLMs.

%% file: text/_1.conclusion.tex
\section{Conclusion}

The label bias of LLMs severely hinders their reliability. We considered different approaches for quantifying this bias. Through extensive experiments with ten LLMs across 279 classification tasks, we found that substantial amounts of label bias exist in LLMs. Moreover, we showed this bias persists as LLMs increase in scale, are instruction-tuned, are provided in-context examples, and even when they are calibrated against such bias. 
We proposed a novel calibration method, which outperforms existing calibration approaches and reduces label bias dramatically. 
Our results highlight the need to better estimate and mitigate LLM biases.

%% file: text/_1.limitations.tex
\section*{Limitations}

\paragraph{Model sizes} Although we experiment with models of several sizes, the models we use are all in the 7B--40B range. We chose not to include relatively small models as these often exhibit poor performance in prompt-based settings. While recent efforts have released better and more efficient models, we leave those for future work. We chose not to experiment with very large LLMs such as Llama 70B due to limitations in computational resources, and as many of them (e.g., GPT-4) are closed~\citep{rogers2023closed}. Therefore, the extent to which our findings apply to such models is unclear. 

\paragraph{Prompt format} Our evaluations are performed on a large and diverse set of tasks extracted from \superni. Still, all tasks contain similar prefixes before introducing instructions, demonstrations and task inputs. Furthermore, each task only has one human-written instruction. We leave experimentation with more varied formats and examination of bias across different instruction phrasings to future work.

\paragraph{Evaluating multilingual tasks} To build our evaluation suite, we extracted tasks from \superni, focusing only on English tasks. We leave analysis on label bias for multilingual tasks to future work.

%% file: text/_1.acknowledgements.tex
\section*{Acknowledgements}

This work was supported in part by the Israel Science Foundation (grant no. 2045/21).

%% file: text/_1.appendix.tex
\appendix

\section{Experimental Setting}
\label{app_experimental_setting}
Our implementation and pretrained model checkpoints use the Huggingface Transformers library~\cite{wolf-etal-2020-transformers}. 
Our code for model evaluation on \superni is based on the code from \citet{wang2023far}. 

\paragraph{Inference} When running inference, we load all models using bf16, except for Falcon-40B, which we load using 8-bit inference. 
We evaluate models using a maximum sequence length of 1024. When incorporating in-context demonstrations into the prompt, the demonstrations are added one by one until the maximal sequence length is reached, while ensuring enough space remains for the input of the evaluated example. Any remaining demonstrations exceeding this length are excluded from the prompt. Consequently, when evaluating tasks with $k$ demonstrations, the contexts for tasks with very long inputs may contain fewer than $k$ demonstrations. In our experiment detailed in \subsecref{level_of_imbalance}, which investigates the impact of label imbalance in the demonstrations set on label bias, we use a sequence length of 2048 and only analyze results for tasks where the prompt contains precisely $k$ demonstrations, excluding other instances from our reported findings.

\paragraph{Compute} We run all experiments on Quadro RTX 6000 (24GB) and RTX A6000 (48GB) GPUs, except for Falcon-40B experiments, which we run on A100 GPUs. 
Average inference run-times on our entire evaluation suite is 18 hours for 7B models, 24 hours for 13B models, and 24 hours for 40B models.
Running LoRA fine-tuning along with inference for 7B models takes 26 hours.
Computing calibration parameters, including running inference on inputs required for calibration, takes around 30 minutes to 2 hours for each model, depending on the method used.

\paragraph{LoRA hyperparameters}
\label{app_lora}
We use all of the hyperparamets used by \citet{dettmers2023qlora} when fine-tuning on \superni, except for using bf16 training instead of 8-bit, a warm-up rate of 0.0, and 5 training epochs. Specifically, we use a learning rate of $0.002$, LoRA $r=64$ and LoRA $\alpha=16$.

\section{Evaluation Suite}
\label{app_superni}

\input{tables/superni/categories}

\input{tables/superni/num_labels}

\input{tables/superni/frequent_labels}

We evaluate models on a subset of 279 tasks from the \superni benchmark~\cite{wang-etal-2022-super}. We use up to 1000 evaluation examples for each task. Altogether, our evaluation set consists of 264,176 examples.

We detail the categories of the selected tasks along with the number of tasks corresponding to each category in~\tabref{superni_task_categories}. We also report the distribution of the number of labels across tasks in~\tabref{superni_num_labels}, as well as the 20 most frequent labels in~\tabref{superni_frequent_labels}.

\section{Supplementary Results}
\label{app_results}

\subsection{Performance and Label Bias}
\label{app_before_mitigation_results}

We provide additional results for the performance and label bias of models~(\subsecref{llms_are_biased}) for Mistral~(\figref{mistral_before_mitigation}) and Falcon~(\figref{falcon_before_mitigation}) models.

\subsection{Bias Mitigation Methods}
\label{app_mitigation_results}
We present additional results for the impact of bias mitigation methods~(\subsecref{effects_of_mitigation}) for Mistral~(\figref{mitigation_methods_mistral}) and Falcon~(\figref{mitigation_methods_falcon}) models.

\subsection{Semantically Equivalent Labels}
\label{app_sem_eq_labels}
We present additional results for the analysis on label bias for tasks with semantically equivalent labels~(\subsecref{sem_eq_labels}) for Mistral~(\figref{sem_eq_labels_Mistral}) and Falcon~(\figref{sem_eq_labels_Falcon}) models.

\subsection{Choice of Demonstrations}
\label{app_bias_sensitivity}

We present additional results  for the analysis on the sensitivity of label bias to the choice of in-context examples~(\subsecref{bias_sensitivity}). We report separate results for each model: Llama-2 7B chat~(\tabref{llama2-7b-chat_icl_set_choice}), Llama-2 13B base~(\tabref{llama2-13b-base_icl_set_choice}), Llama-2 13B chat~(\tabref{llama2-13b-chat_icl_set_choice}), Mistral 7B base~(\tabref{mistral-7b-base_icl_set_choice}), and Mistral 7B instruct~(\tabref{mistral-7b-instruct_icl_set_choice}).

\input{figures/sem_eq_labels/mistral}
\input{figures/sem_eq_labels/falcon}

\input{tables/llama2-7b_chat_icl_sets}

\input{tables/llama2-13_base_icl_sets}

\input{tables/llama2-13b_chat_icl_sets}

\input{tables/mistral-7b_base_icl_sets}

\input{tables/mistral-7b_instruct_icl_sets}

\input{figures/before_mitigation/mistral}

\input{figures/before_mitigation/falcon}

\input{figures/mitigation_methods/mistral}

\input{figures/mitigation_methods/falcon}

%% file: tables/superni/categories.tex
\begin{table}[!b]
\centering
\resizebox{0.9\columnwidth}{!}{%
\begin{tabular}{@{}ccc@{}}
\toprule
\textbf{Task Category}                  & \textbf{\# of Tasks} & \textbf{\# of Instances} \\ \midrule
Sentiment Analysis             & 39         & 37748          \\
Text Categorization            & 30         & 27652          \\
Toxic Language Detection       & 25         & 24114          \\
Commonsense Classification     & 23         & 22239          \\
Textual Entailment             & 15         & 14613          \\
Question Answering             & 13         & 12380          \\
Answerability Classification   & 12         & 11286          \\
Text Matching                  & 11         & 10807          \\
Question Understanding         & 8          & 7730           \\
Text Completion                & 7          & 7000           \\
Speaker Identification         & 6          & 4739           \\
Ethics Classification          & 6          & 5501           \\
Text Quality Evaluation        & 6          & 6000           \\
Dialogue Act Recognition       & 6          & 5401           \\
Stereotype Detection           & 6          & 5627           \\
Cause Effect Classification    & 5          & 4200           \\
Word Relation Classification   & 5          & 4680           \\
Gender Classification          & 5          & 5000           \\
Negotiation Strategy Detection & 5          & 4150           \\
Coherence Classification       & 5          & 5000           \\
Answer Verification            & 4          & 4000           \\
Information Extraction         & 4          & 4000           \\
Dialogue State Tracking        & 3          & 2855           \\
Coreference Resolution         & 3          & 3000           \\
Linguistic Probing             & 2          & 1808           \\
Pos Tagging                    & 2          & 2000           \\
Irony Detection                & 2          & 1933           \\
Word Semantics                 & 2          & 1210           \\
Text to Code                   & 2          & 2000           \\
Intent Identification          & 2          & 2000           \\
Section Classification         & 2          & 2000           \\
Tasks With Unique Categories                         & 13         & 11503          \\ 
\midrule

Total & 279 & 264,176

\\ \bottomrule
\end{tabular}
}

\caption{
Categories of tasks included in our evaluation suite, based on \superni, along with the number of tasks per category and the total number of instances used for evaluating models.
}

\label{tab:superni_task_categories}
\end{table}

%% file: tables/superni/num_labels.tex
\begin{table}[tb]
\centering
\resizebox{0.8\columnwidth}{!}{%
\begin{tabular}{@{}ccc@{}}
\toprule
\textbf{Answer Choices}                  & \textbf{Number of Tasks} \\ \midrule
2             & 39                \\
3            & 30            \\
4       & 25                 \\
5     & 23              \\ 
6-9     & 14              \\ 
10+     & 8              \\
\bottomrule
\end{tabular}
}

\caption{
Distribution of the number of labels across tasks in our evaluation suite.
}

\label{tab:superni_num_labels}
\end{table}

%% file: tables/superni/frequent_labels.tex
\begin{table}[tb]
\centering
\resizebox{0.8\columnwidth}{!}{%
\begin{tabular}{@{}cc|cc@{}}
\textbf{Label} & \textbf{Freq.} & \textbf{Label}  & \textbf{Freq.} \\ \midrule

no             & 76                      & 3               & 14                      \\
yes            & 75                      & negative        & 13                      \\
1              & 31                      & 4               & 10                      \\
true           & 20                      & c               & 9                       \\
false          & 20                      & 5               & 9                       \\
b              & 19                      & neutral         & 8                       \\
a              & 19                      & d               & 7                       \\
2              & 18                      & anti-stereotype & 5                       \\
0              & 17                      & stereotype      & 5                       \\
positive       & 14                      & pos             & 5                       \\ \bottomrule
\end{tabular}
}

\caption{
The 20 most frequent labels in our evaluation suite and the number of tasks they appear in.
}

\label{tab:superni_frequent_labels}
\end{table}

%% file: figures/sem_eq_labels/mistral.tex
\begin{figure}[!tb]
    \centering
    \captionsetup[subfigure]
    {font=small,labelfont=small} 
    
    \begin{subfigure}[tb]{0.9\columnwidth}
        \centering
        
        \includegraphics[width=\columnwidth]{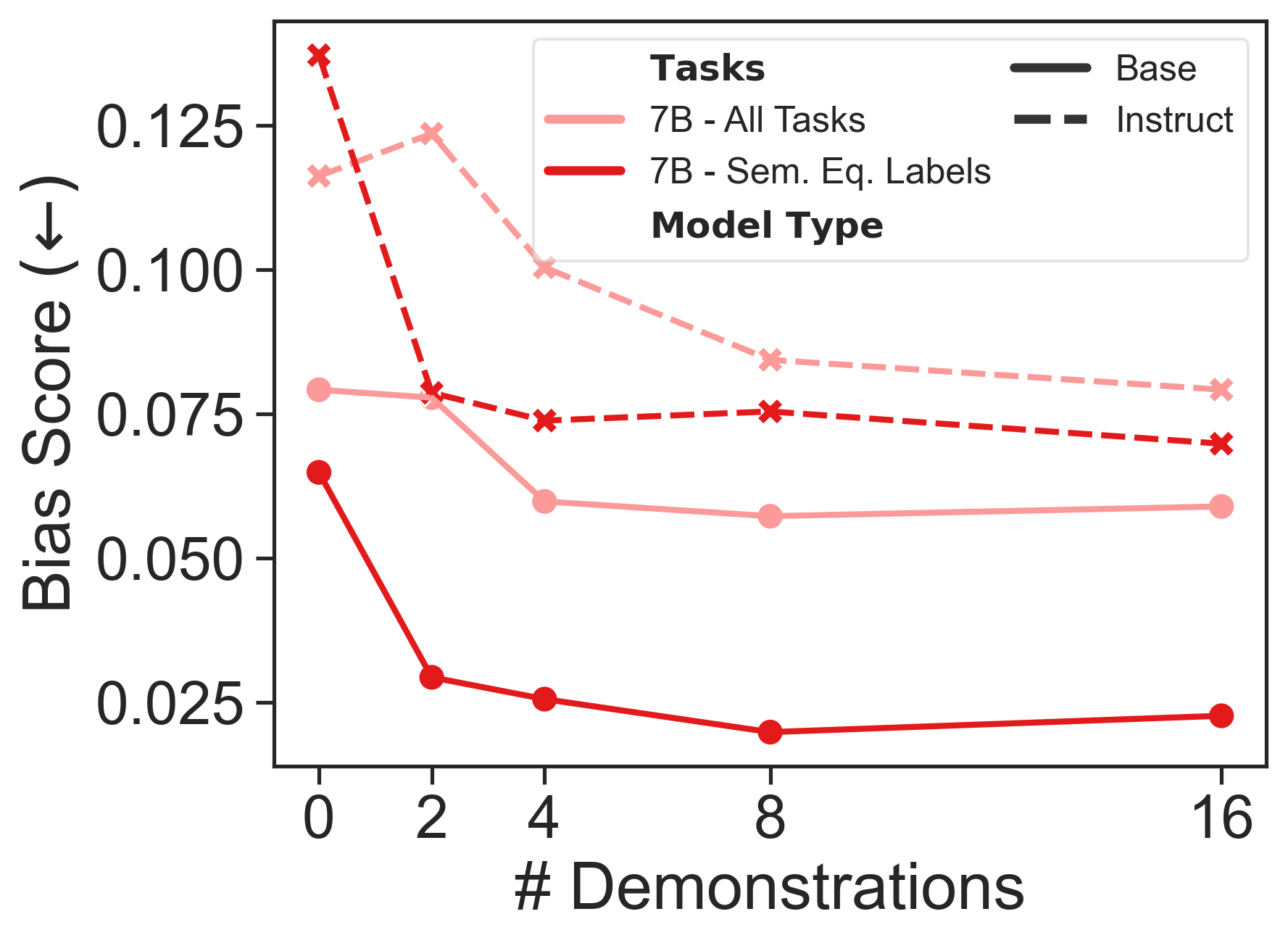}

        \subcaption{\bias}
    \end{subfigure}%
 
    \begin{subfigure}[tb]{0.9\columnwidth}
        \centering
        
        \includegraphics[width=\columnwidth]{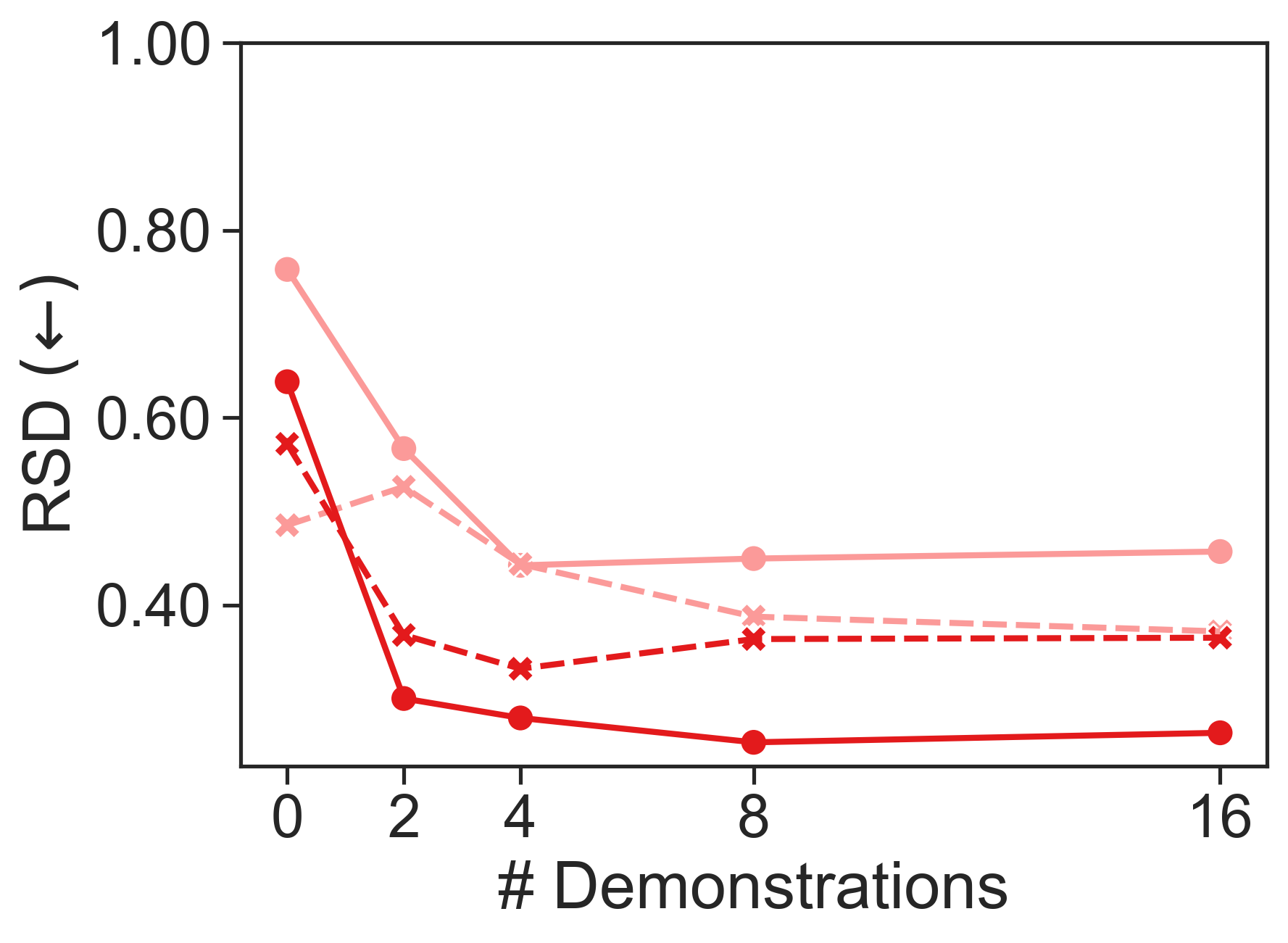}
        
        \subcaption{\rsd}
    \end{subfigure}%
    ~
    \caption{Label bias metrics for Mistral 7B models when evaluated on all tasks in our evaluation suite (\emph{All}) vs.~a subset of tasks with semantically equivalent labels (\emph{Sem.~Eq.~Labels}).
    }
    
\label{fig:sem_eq_labels_Mistral}
\end{figure}

%% file: figures/sem_eq_labels/falcon.tex
\begin{figure}[t]
    \centering
    \captionsetup[subfigure]
    {font=small,labelfont=small} 
    
    \begin{subfigure}[t]{0.9\columnwidth}
        \centering
        
        \includegraphics[width=\columnwidth]{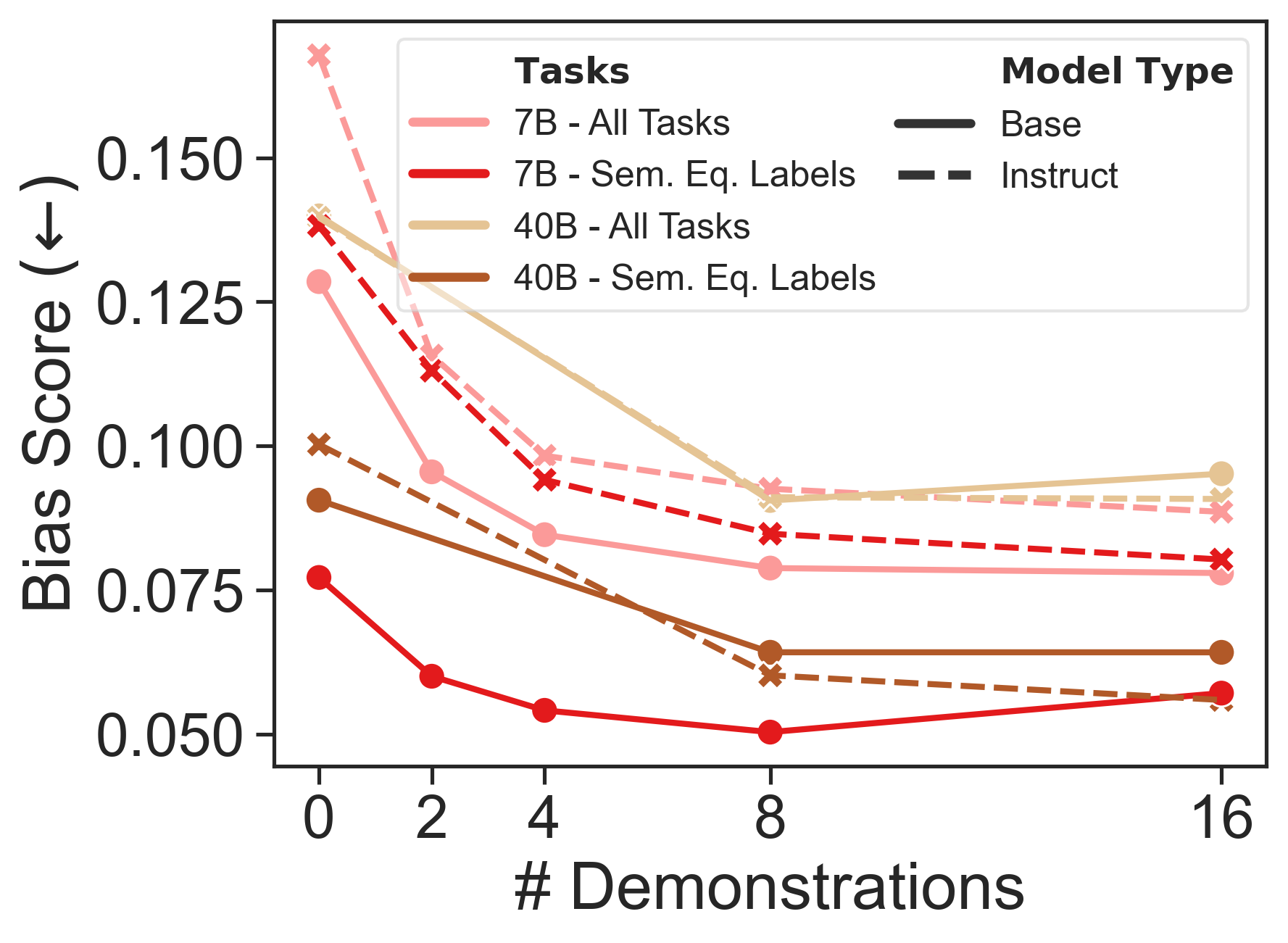}

        \subcaption{\bias}
    \end{subfigure}%
 
    \begin{subfigure}[t]{0.9\columnwidth}
        \centering
        
        \includegraphics[width=\columnwidth]{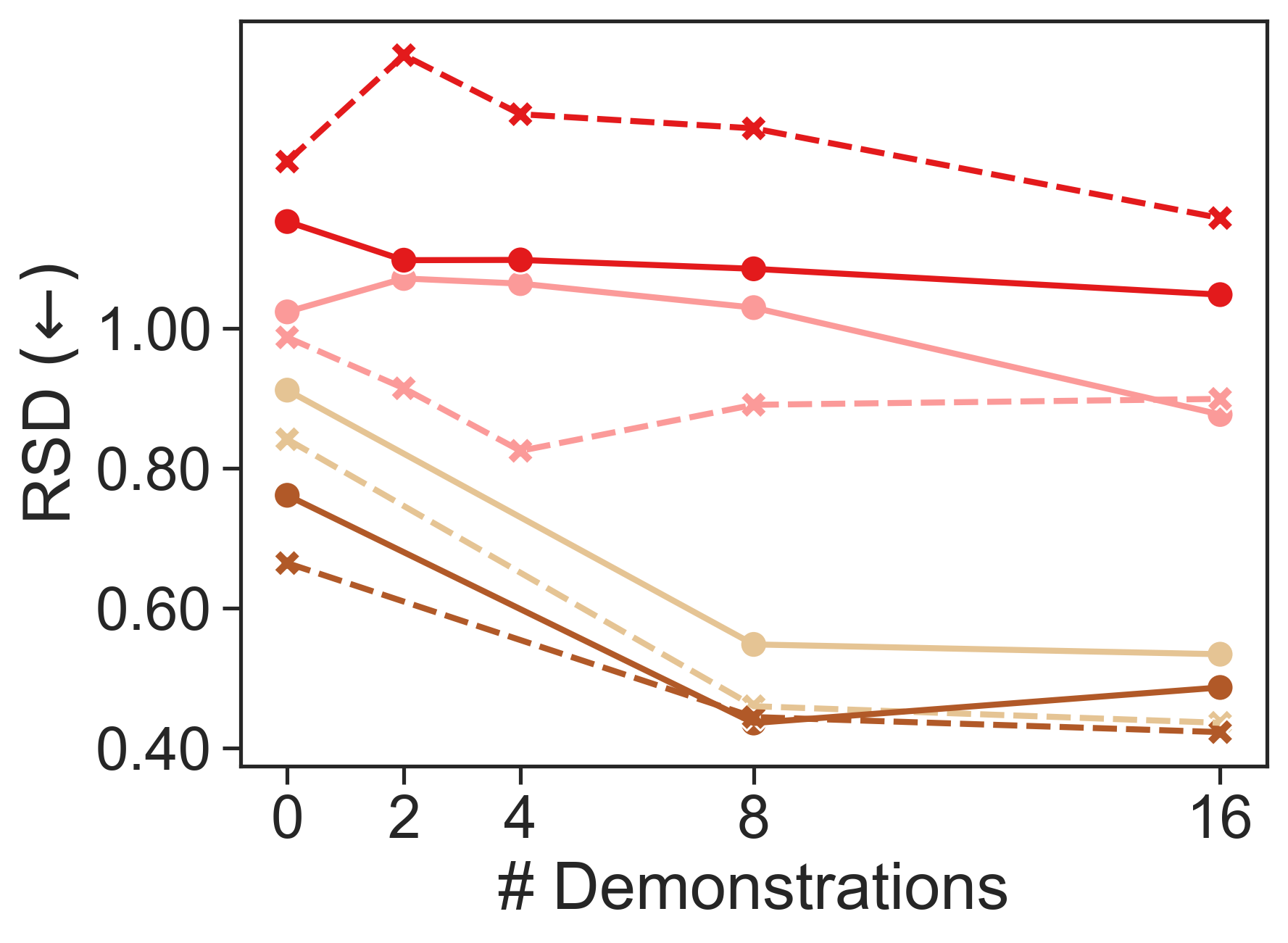}
        
        \subcaption{\rsd}
    \end{subfigure}%
    ~
    \caption{Label bias metrics for Falcon (7B/40B) models when evaluated on all tasks in our evaluation suite (\emph{All}) vs.~a subset of tasks with semantically equivalent labels (\emph{Sem.~Eq.~Labels}).
    }
    
\label{fig:sem_eq_labels_Falcon}
\end{figure}

%% file: tables/llama2-7b_chat_icl_sets.tex
\begin{table}[!tb]
\centering
\resizebox{\columnwidth}{!}{%
\begin{tabular}{@{}cccc@{}}
\toprule
\textbf{Choice of Demonstrations} & \textbf{F1} ($\uparrow$)           & \textbf{\rsd} ($\downarrow$)         & \textbf{\bias} ($\downarrow$)         \\ \midrule
Mean (SD)                     & 0.571 ($\pm$ 0.059) & 0.417 ($\pm$ 0.178) & 0.141 ($\pm$ 0.061) \\ \midrule
Best Performance              & 0.636        & 0.267        & 0.095         \\
Median Performance              & 0.577        & 0.401        & 0.137         \\
Worst Performance             & 0.494        & 0.613        & 0.197         \\ \midrule
Least Bias -- \textsl{by} \rsd             & 0.619        & 0.216         & 0.084         \\
Least Bias -- \textsl{by} \bias            & 0.614        & 0.25        & 0.07         \\
Most Bias -- \textsl{by} \rsd              & 0.503        & 0.645        & 0.205         \\
Most Bias -- \textsl{by} \bias             & 0.511        & 0.613        & 0.22         \\ \bottomrule
\end{tabular}

}

\caption{
Results of Llama-2 7B chat model when prompted with 5 different sets of demonstration on our evaluation suite. We employ oracles to aggregate and compute cross-task results when choosing a specific set of demonstrations for each task.  
}
\label{tab:llama2-7b-chat_icl_set_choice}
\end{table}

%% file: tables/llama2-13_base_icl_sets.tex
\begin{table}[!tb]
\centering
\resizebox{\columnwidth}{!}{%
\begin{tabular}{@{}cccc@{}}
\toprule
\textbf{Choice of Demonstrations} & \textbf{F1} ($\uparrow$)           & \textbf{\rsd} ($\downarrow$)         & \textbf{\bias} ($\downarrow$)         \\ \midrule
Mean (SD)                     & 0.54 ($\pm$ 0.069) & 0.546 ($\pm$ 0.205) & 0.088 ($\pm$ 0.031) \\ \midrule
Best Performance              & 0.618        & 0.352        & 0.065         \\
Median Performance              & 0.544        & 0.522        & 0.084         \\
Worst Performance             & 0.452        & 0.782        & 0.114         \\ \midrule
Least Bias -- \textsl{by} \rsd             & 0.605        & 0.314         & 0.062         \\
Least Bias -- \textsl{by} \bias            & 0.592        & 0.369        & 0.052         \\
Most Bias -- \textsl{by} \rsd              & 0.457        & 0.806        & 0.118         \\
Most Bias -- \textsl{by} \bias             & 0.475        & 0.747        & 0.128         \\ \bottomrule
\end{tabular}

}

\caption{
Results of Llama-2 13B base model. 
}
\label{tab:llama2-13b-base_icl_set_choice}
\end{table}

%% file: tables/llama2-13b_chat_icl_sets.tex
\begin{table}[t]
\centering
\resizebox{\columnwidth}{!}{%
\begin{tabular}{@{}cccc@{}}
\toprule
\textbf{Choice of Demonstrations} & \textbf{F1} ($\uparrow$)           & \textbf{\rsd} ($\downarrow$)         & \textbf{\bias} ($\downarrow$)         \\ \midrule
Mean (SD)                     & 0.592 ($\pm$ 0.058) & 0.397 ($\pm$ 0.173) & 0.134 ($\pm$ 0.058) \\ \midrule
Best Performance              & 0.656        & 0.241        & 0.092         \\
Median Performance              & 0.597        & 0.383        & 0.13         \\
Worst Performance             & 0.517        & 0.584        & 0.185         \\ \midrule
Least Bias -- \textsl{by} \rsd             & 0.643        & 0.201         & 0.085         \\
Least Bias -- \textsl{by} \bias            & 0.632        & 0.251        & 0.067         \\
Most Bias -- \textsl{by} \rsd              & 0.523        & 0.622        & 0.194         \\
Most Bias -- \textsl{by} \bias             & 0.534        & 0.58        & 0.21         \\ \bottomrule
\end{tabular}

}

\caption{
Results of Llama-2 13B chat model.
}
\label{tab:llama2-13b-chat_icl_set_choice}
\end{table}

%% file: tables/mistral-7b_base_icl_sets.tex
\begin{table}[t]
\centering
\resizebox{\columnwidth}{!}{%
\begin{tabular}{@{}cccc@{}}
\toprule
\textbf{Choice of Demonstrations} & \textbf{F1} ($\uparrow$)           & \textbf{\rsd} ($\downarrow$)         & \textbf{\bias} ($\downarrow$)         \\ \midrule
Mean (SD)                     & 0.601 ($\pm$ 0.092) & 0.432 ($\pm$ 0.239) & 0.064 ($\pm$ 0.034) \\ \midrule
Best Performance              & 0.692        & 0.225        & 0.057         \\
Median Performance              & 0.616        & 0.387        & 0.064         \\
Worst Performance             & 0.47        & 0.747        & 0.064         \\ \midrule
Least Bias -- \textsl{by} \rsd             & 0.68        & 0.196         & 0.055         \\
Least Bias -- \textsl{by} \bias            & 0.579        & 0.484        & 0.021         \\
Most Bias -- \textsl{by} \rsd              & 0.477        & 0.77        & 0.067         \\
Most Bias -- \textsl{by} \bias             & 0.563        & 0.537        & 0.102         \\ \bottomrule
\end{tabular}

}

\caption{
Results of Mistral 7B base model.
}

\label{tab:mistral-7b-base_icl_set_choice}
\end{table}

%% file: tables/mistral-7b_instruct_icl_sets.tex
\begin{table}[t]
\centering
\resizebox{\columnwidth}{!}{%
\begin{tabular}{@{}cccc@{}}
\toprule
\textbf{Choice of Demonstrations} & \textbf{F1} ($\uparrow$)           & \textbf{\rsd} ($\downarrow$)         & \textbf{\bias} ($\downarrow$)         \\ \midrule
Mean (SD)                     & 0.607 ($\pm$ 0.059) & 0.389 ($\pm$ 0.167) & 0.085 ($\pm$ 0.035) \\ \midrule
Best Performance              & 0.672        & 0.232        & 0.06         \\
Median Performance              & 0.613        & 0.368        & 0.08         \\
Worst Performance             & 0.53        & 0.585        & 0.115         \\ \midrule
Least Bias -- \textsl{by} \rsd             &  0.663        & 0.202         & 0.059         \\
Least Bias -- \textsl{by} \bias            & 0.653        & 0.245        & 0.044         \\
Most Bias -- \textsl{by} \rsd              & 0.535        & 0.604        & 0.119         \\
Most Bias -- \textsl{by} \bias             & 0.553        & 0.545        & 0.131         \\ \bottomrule
\end{tabular}

}

\caption{
Results of Mistral 7B instruct model. 
}

\vspace{51.5mm}

\label{tab:mistral-7b-instruct_icl_set_choice}
\end{table}

%% file: figures/before_mitigation/mistral.tex
\begin{figure*}[b]
    \centering
    \begin{subfigure}[t]{0.31\textwidth}
        \centering
        \includegraphics[width=\textwidth]{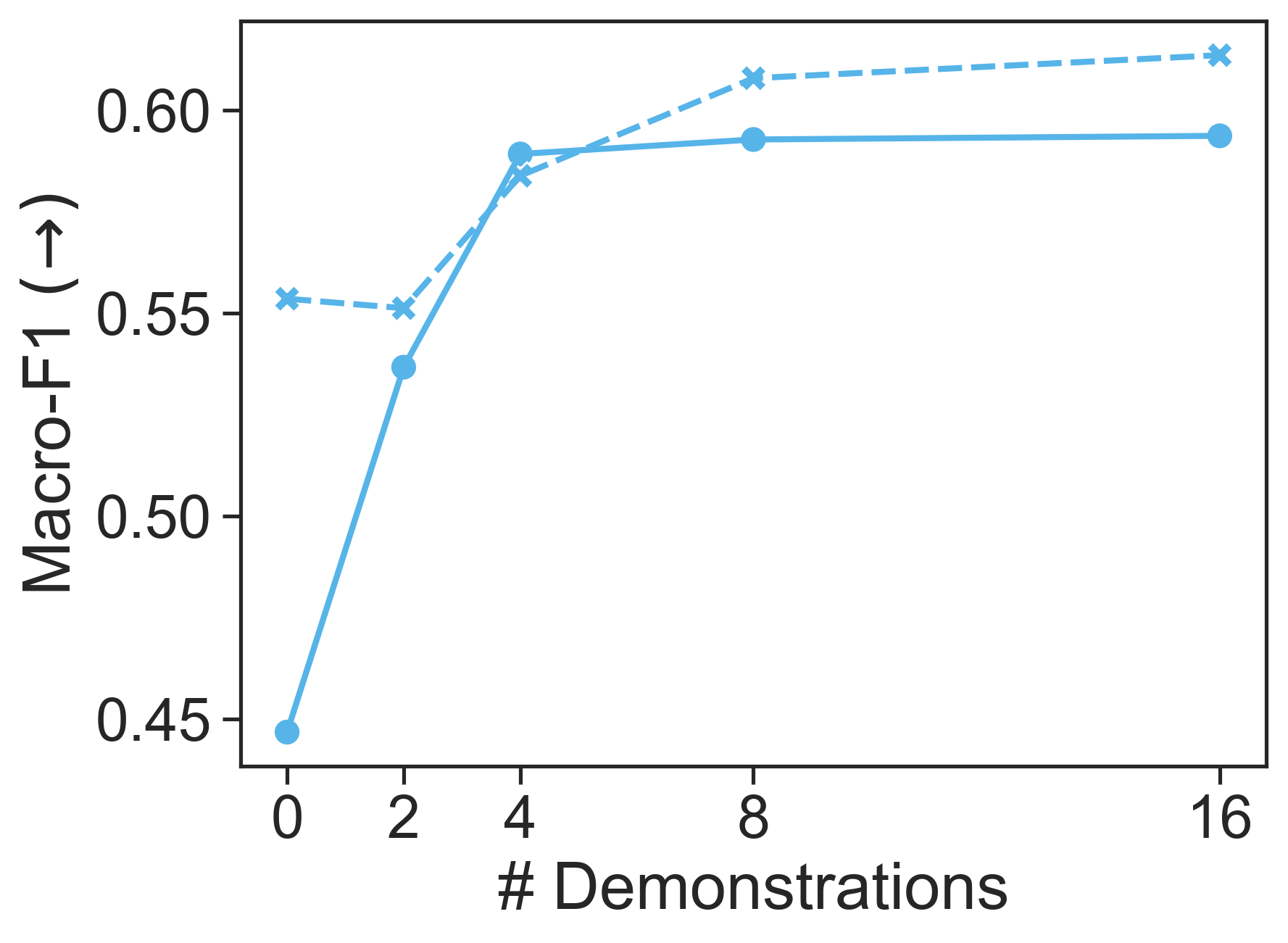}
        \caption{Performance (Macro-F1)}
    \end{subfigure}%
    ~ 
    \begin{subfigure}[t]{0.31\textwidth}
        \centering
        \includegraphics[width=\textwidth]{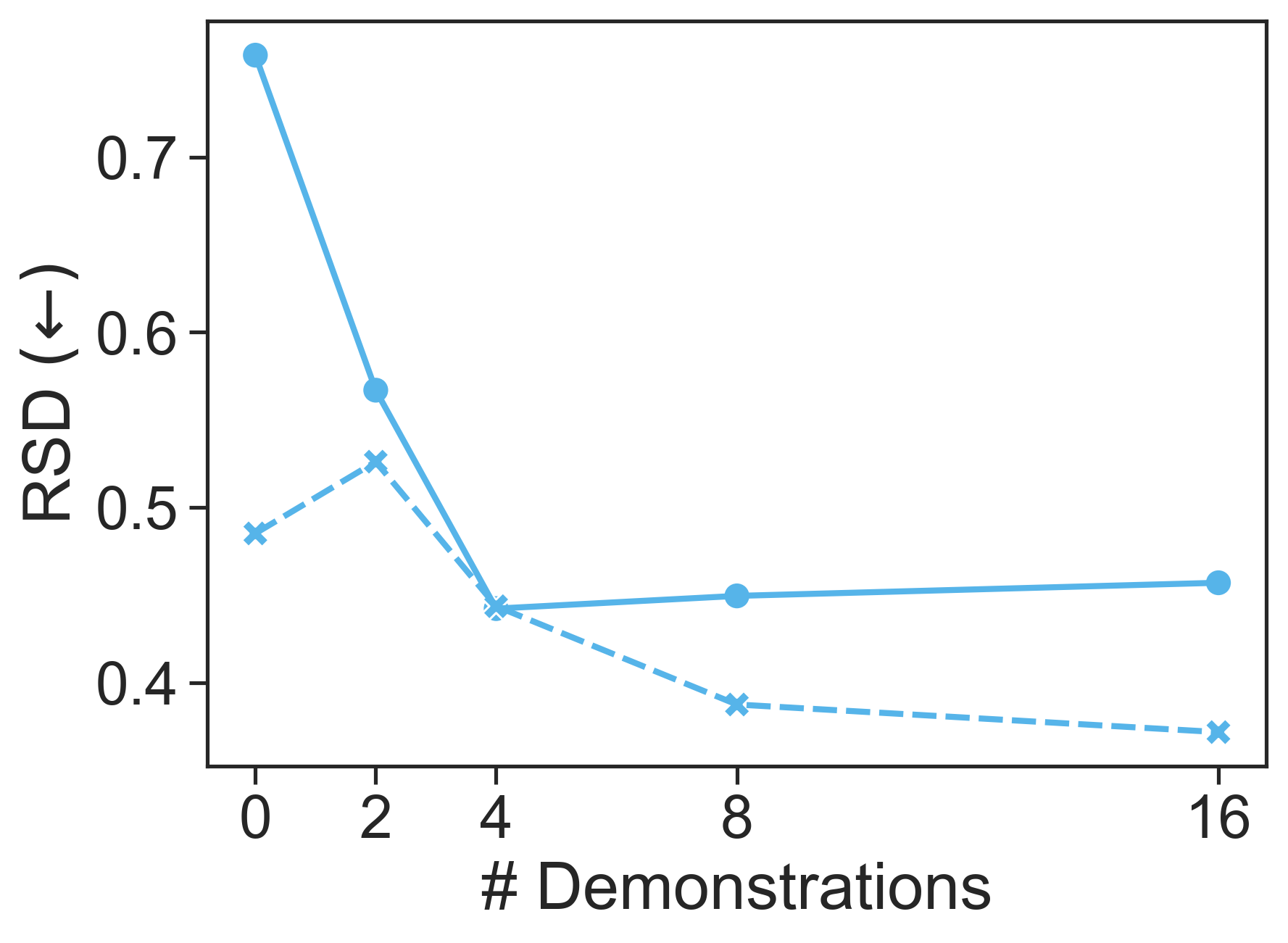}
        \caption{Label bias (\rsd)}
    \end{subfigure}
    ~
    \begin{subfigure}[t]{0.31\textwidth}
        \centering
        \includegraphics[width=\textwidth]{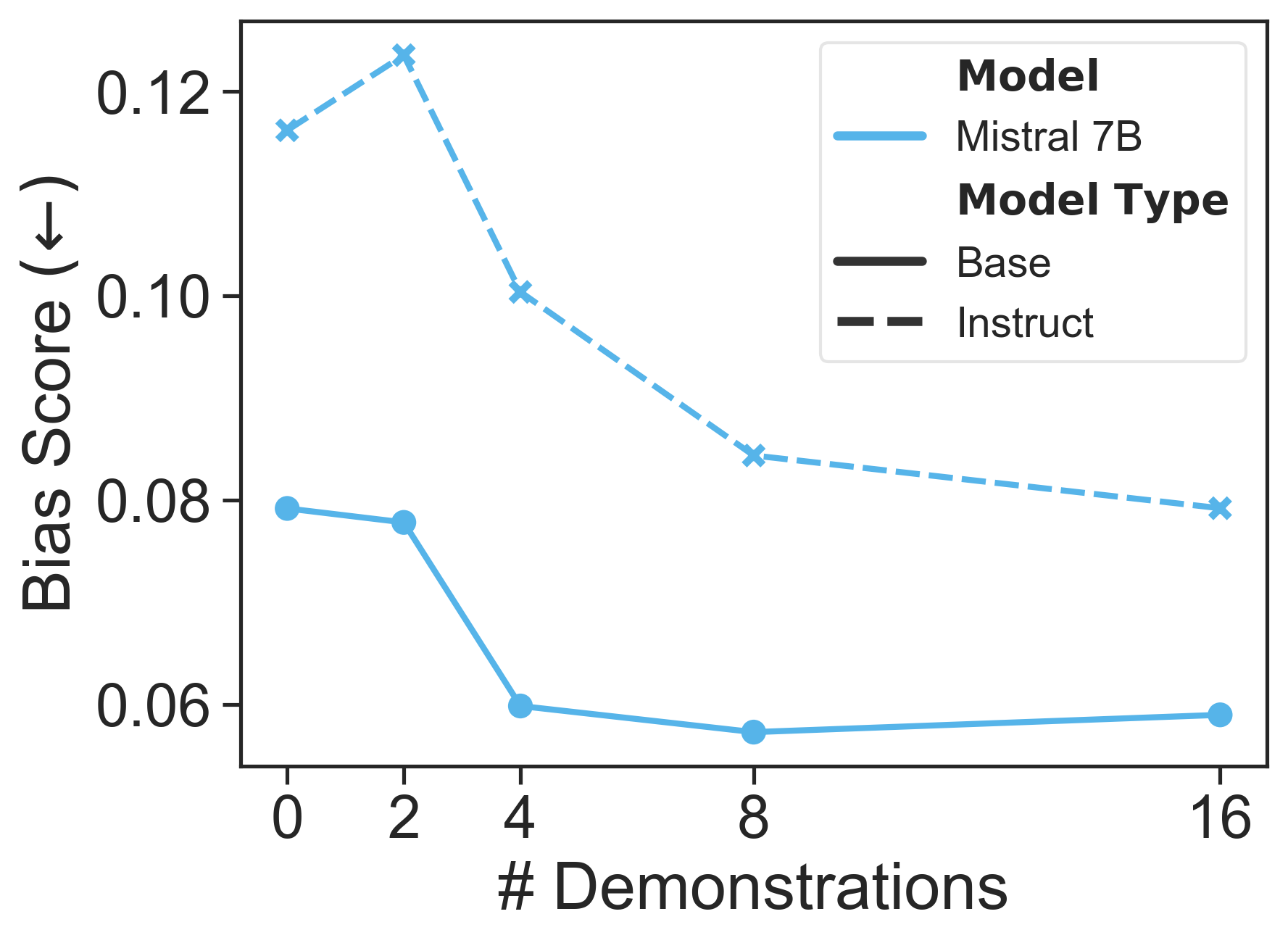}
        \caption{Label bias (\bias)}
    \end{subfigure}%
    
    \caption{Performance and label bias metrics for Mistral 7B  pretrained and instruction-tuned models.}
\label{fig:mistral_before_mitigation}
\end{figure*}

%% file: figures/before_mitigation/falcon.tex
\begin{figure*}[b]
    \centering
    \begin{subfigure}[t]{0.31\textwidth}
        \centering
        \includegraphics[width=\textwidth]{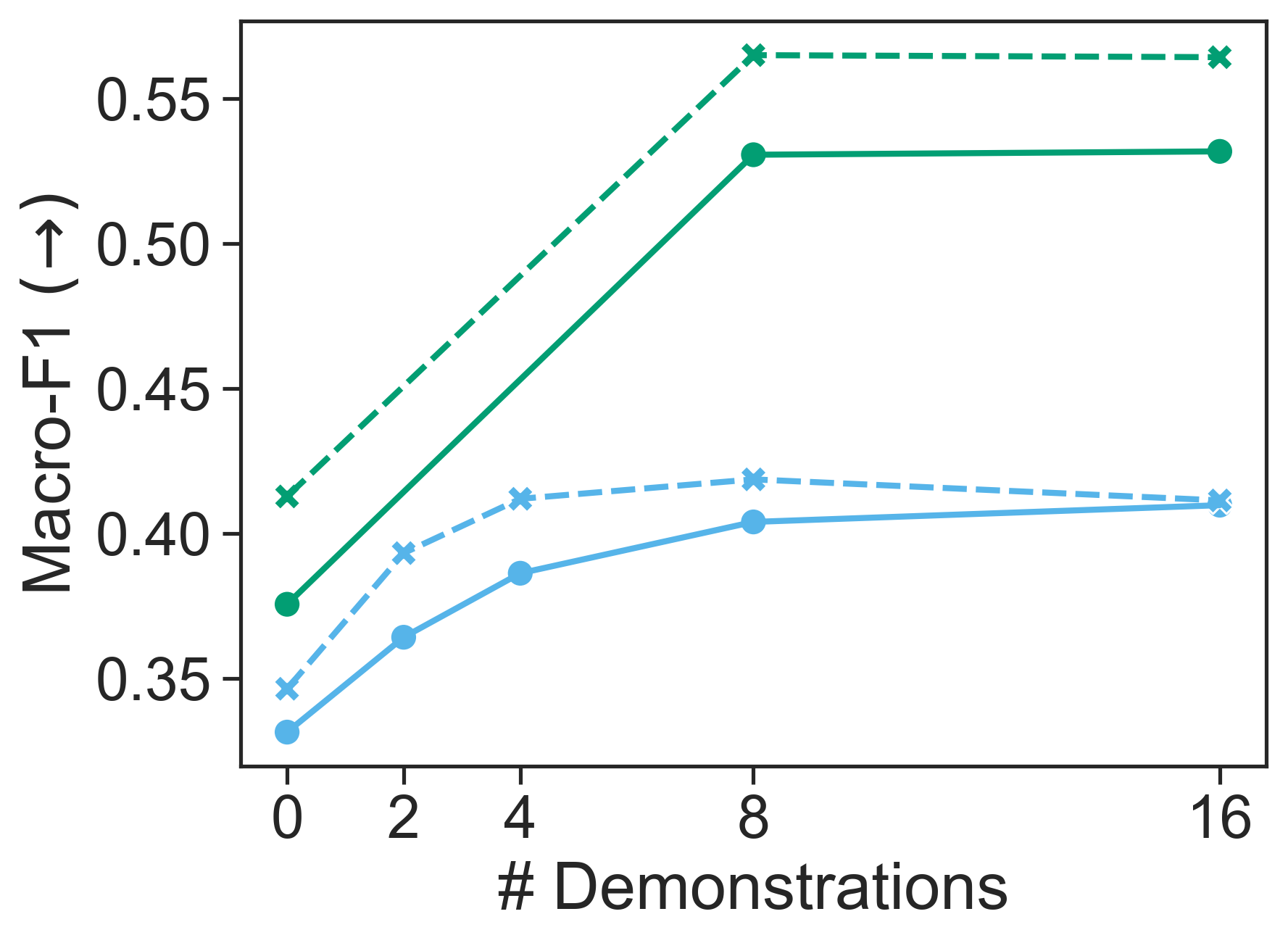}
        \caption{Performance (Macro-F1)}
    \end{subfigure}%
    ~ 
    \begin{subfigure}[t]{0.31\textwidth}
        \centering
        \includegraphics[width=\textwidth]{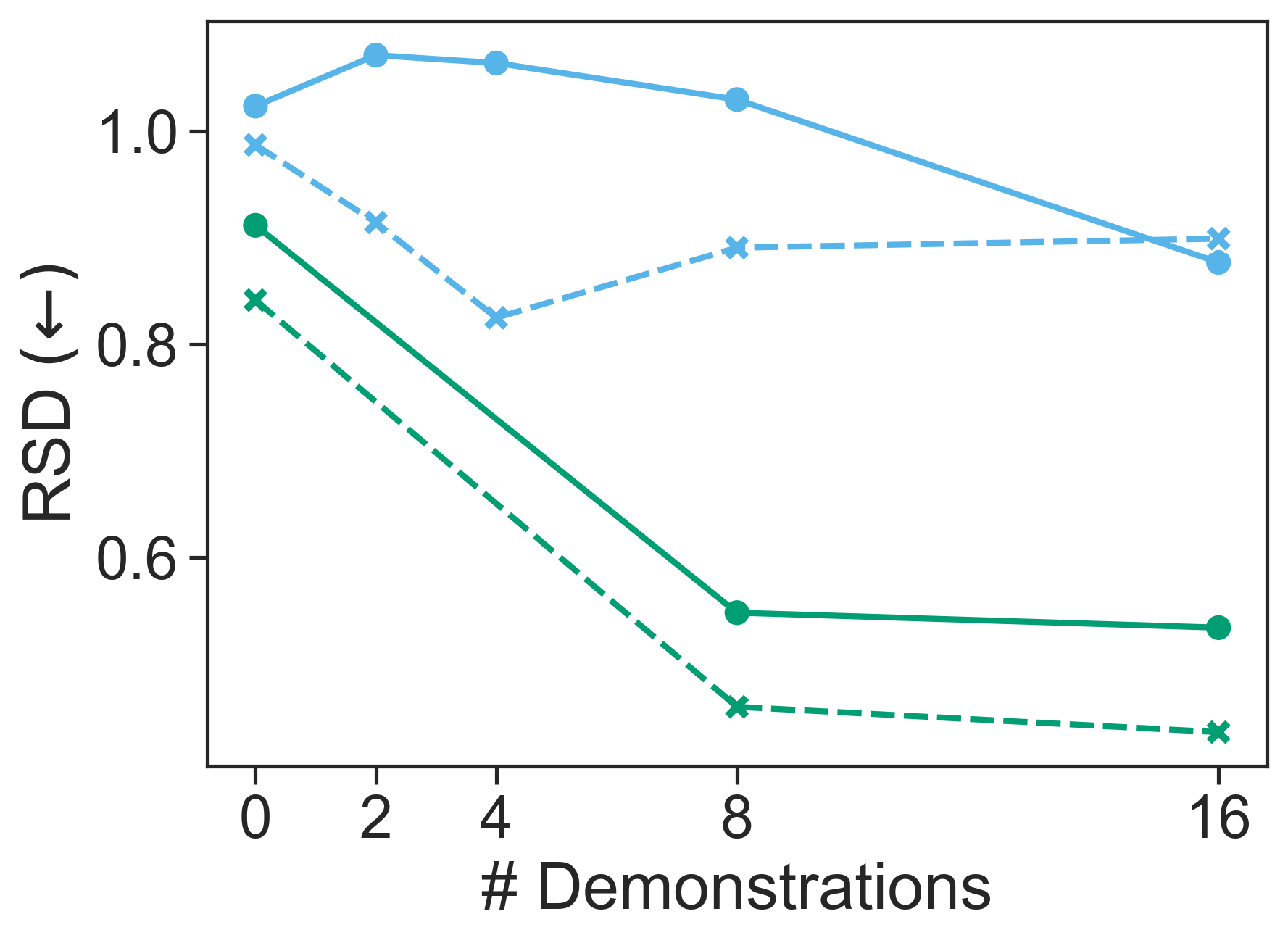}
        \caption{Label bias (\rsd)}
    \end{subfigure}
    ~
    \begin{subfigure}[t]{0.31\textwidth}
        \centering
        \includegraphics[width=\textwidth]{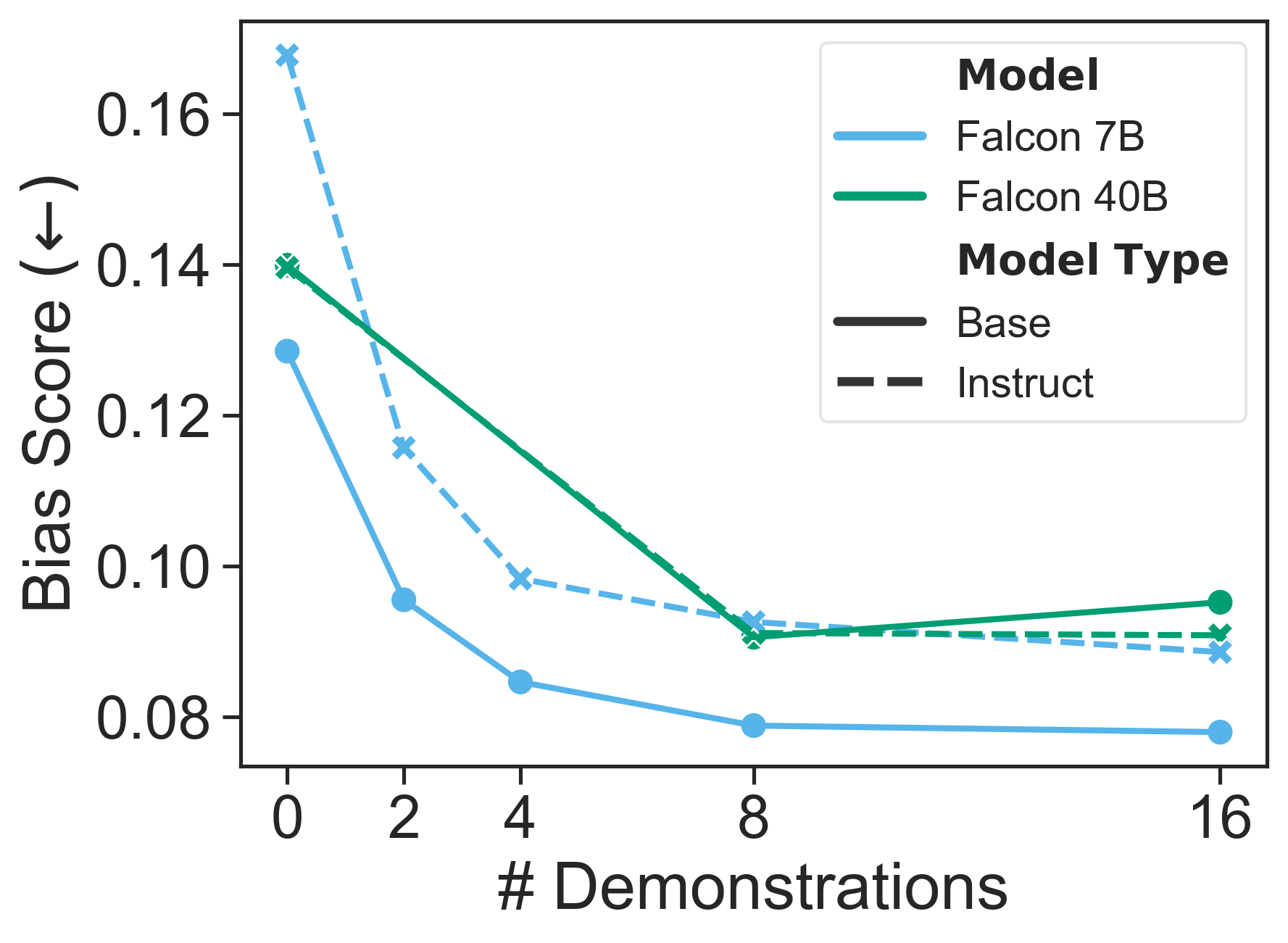}
        \caption{Label bias (\bias)}
    \end{subfigure}%
    
    \caption{Performance and label bias metrics for Falcon  pretrained and instruction-tuned models (7B/40B).}
\label{fig:falcon_before_mitigation}
\end{figure*}

%% file: figures/mitigation_methods/mistral.tex
\begin{figure*}[t]
    \centering
    \captionsetup[subfigure]
    {font=small,labelfont=small} 
    
    \begin{subfigure}[b]{0.03\textwidth}
        \centering
        \mbox{}
        \vfill
        \rotatebox[origin=c]{90}{\small 7B}
        \bigskip
        \vfill
        \vspace{1.7cm}
        \mbox{}
    \end{subfigure}%
    \begin{subfigure}[b]{0.31\textwidth}
        \centering
        
        \includegraphics[width=\textwidth]{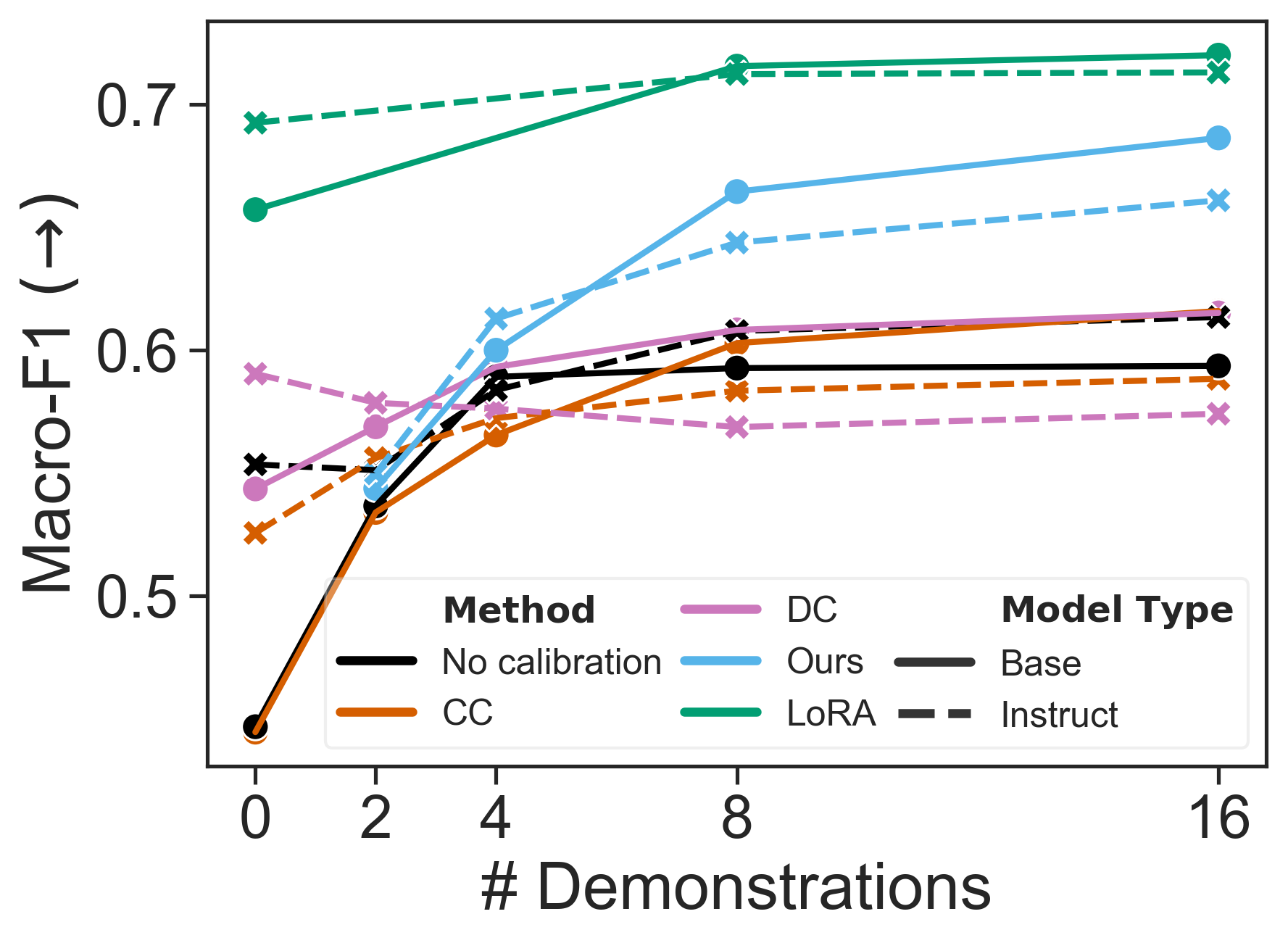}
        
        \subcaption{\hspace{0.5cm}\textsl{Macro-F1}}
        
    \end{subfigure}%
    ~ 
    \begin{subfigure}[b]{0.31\textwidth}
        \centering
        
        \includegraphics[width=\textwidth]{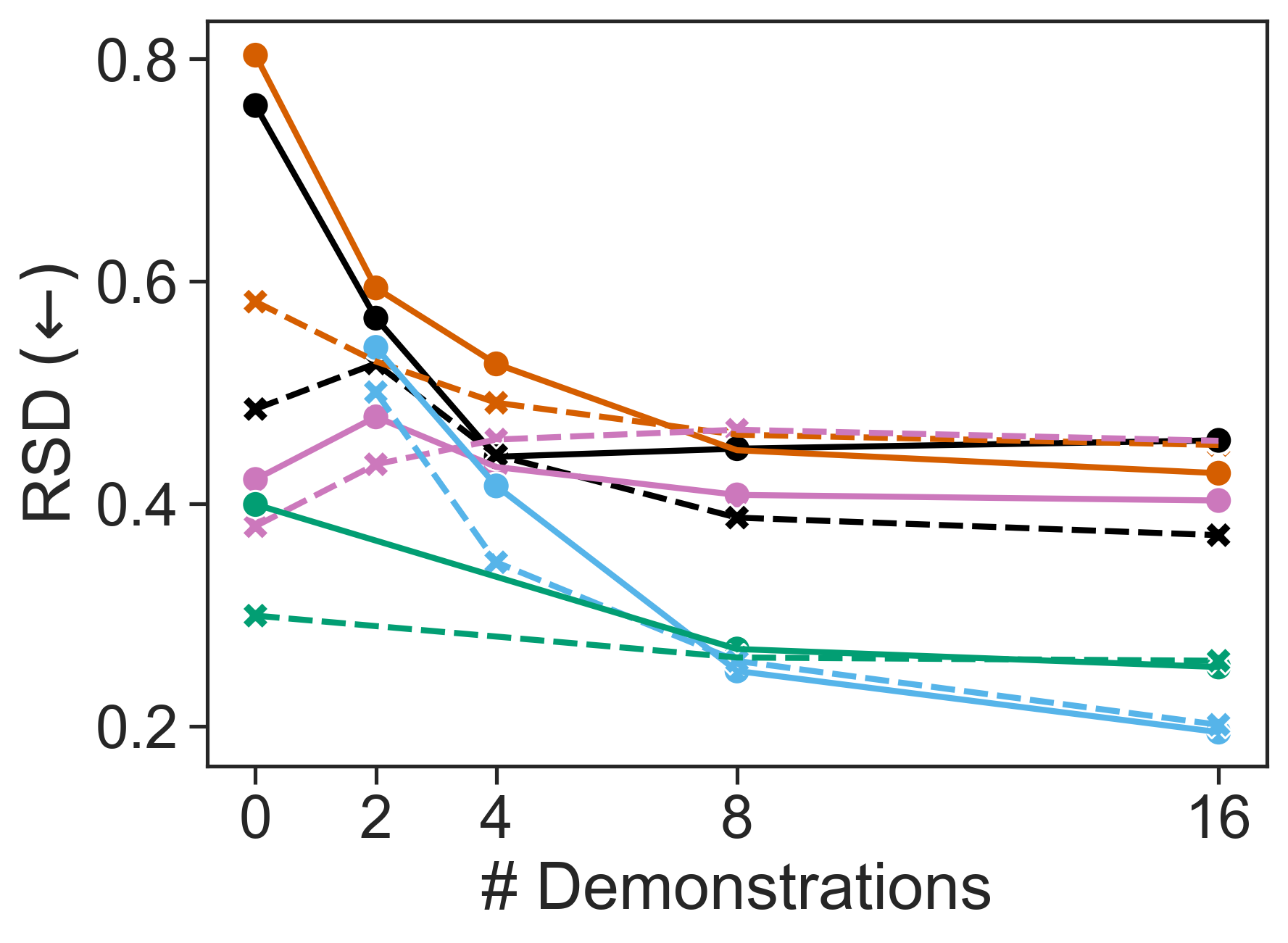}

        \subcaption{\hspace{0.5cm} \textsl{RSD}}

    \end{subfigure}
    ~
    \begin{subfigure}[b]{0.31\textwidth}
        \centering
        
        \includegraphics[width=\textwidth]{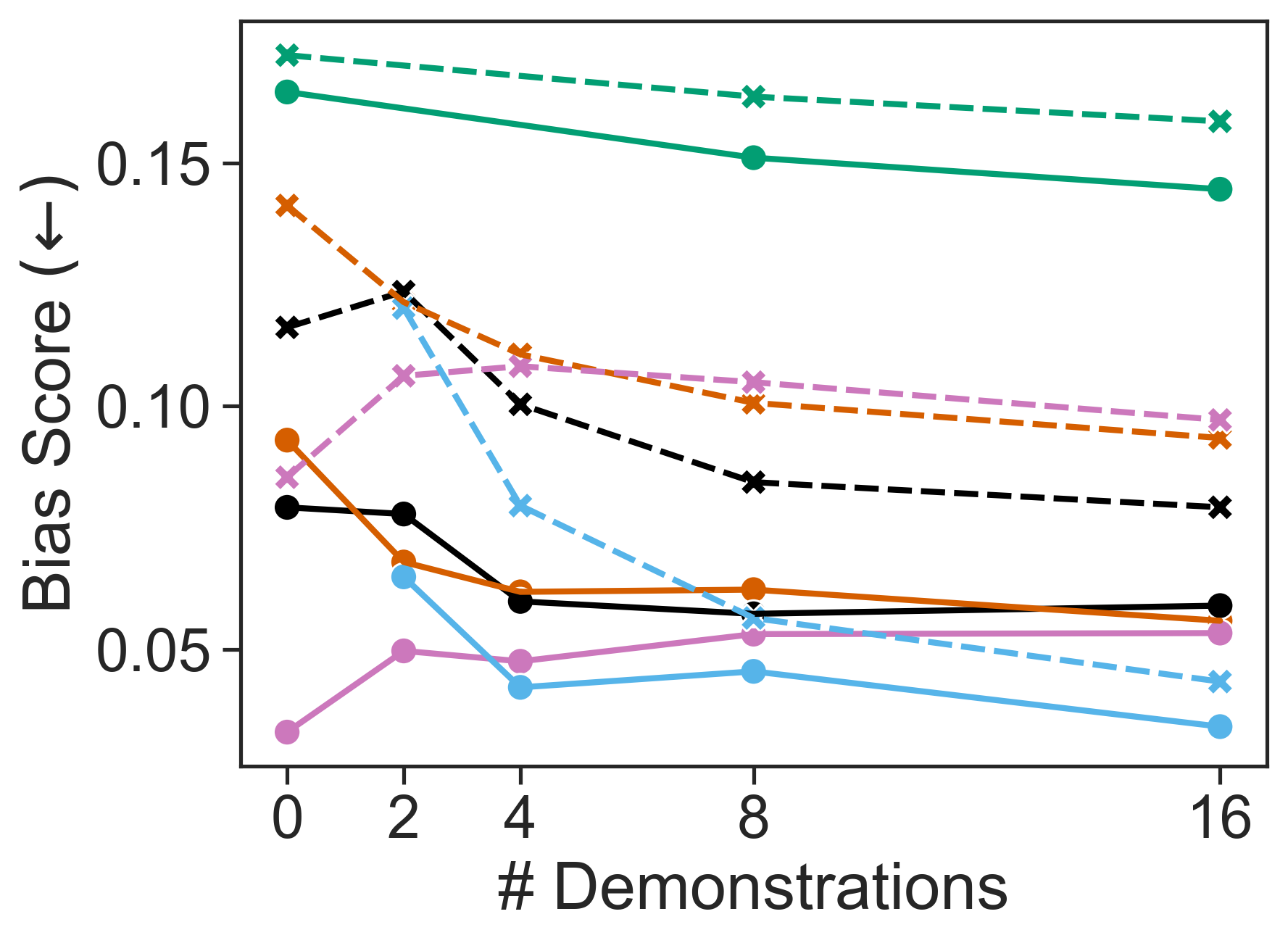}

        \subcaption{\hspace{0.5cm} \textsl{Bias Score}}

    \end{subfigure}%

    \caption{The effect of label bias mitigation methods on performance and bias for Mistral models.}
    
\label{fig:mitigation_methods_mistral}
\end{figure*}

%% file: figures/mitigation_methods/falcon.tex
\begin{figure*}[t]
    \centering
    \captionsetup[subfigure]
    {font=small,labelfont=small} 
    
    \begin{subfigure}[b]{0.03\textwidth}
        \centering
        \mbox{}
        \vfill
        \rotatebox[origin=c]{90}{\small 7B}
        \bigskip
        \vfill
        \vspace{1.1cm}
        \mbox{}
    \end{subfigure}%
    \begin{subfigure}[b]{0.31\textwidth}
        \centering
        
        \includegraphics[width=\textwidth]{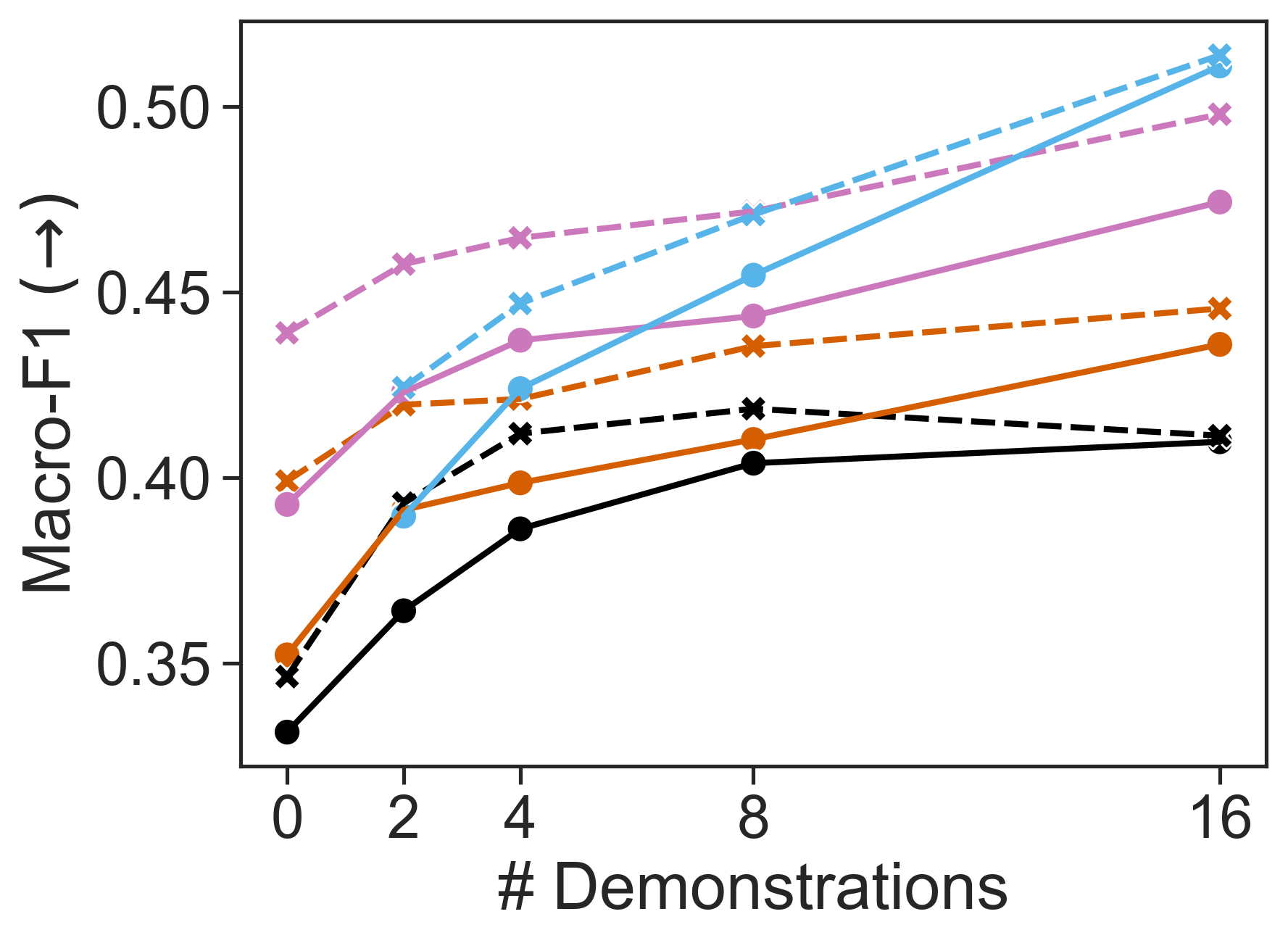}
    \end{subfigure}%
    ~ 
    \begin{subfigure}[b]{0.31\textwidth}
        \centering
        
        \includegraphics[width=\textwidth]{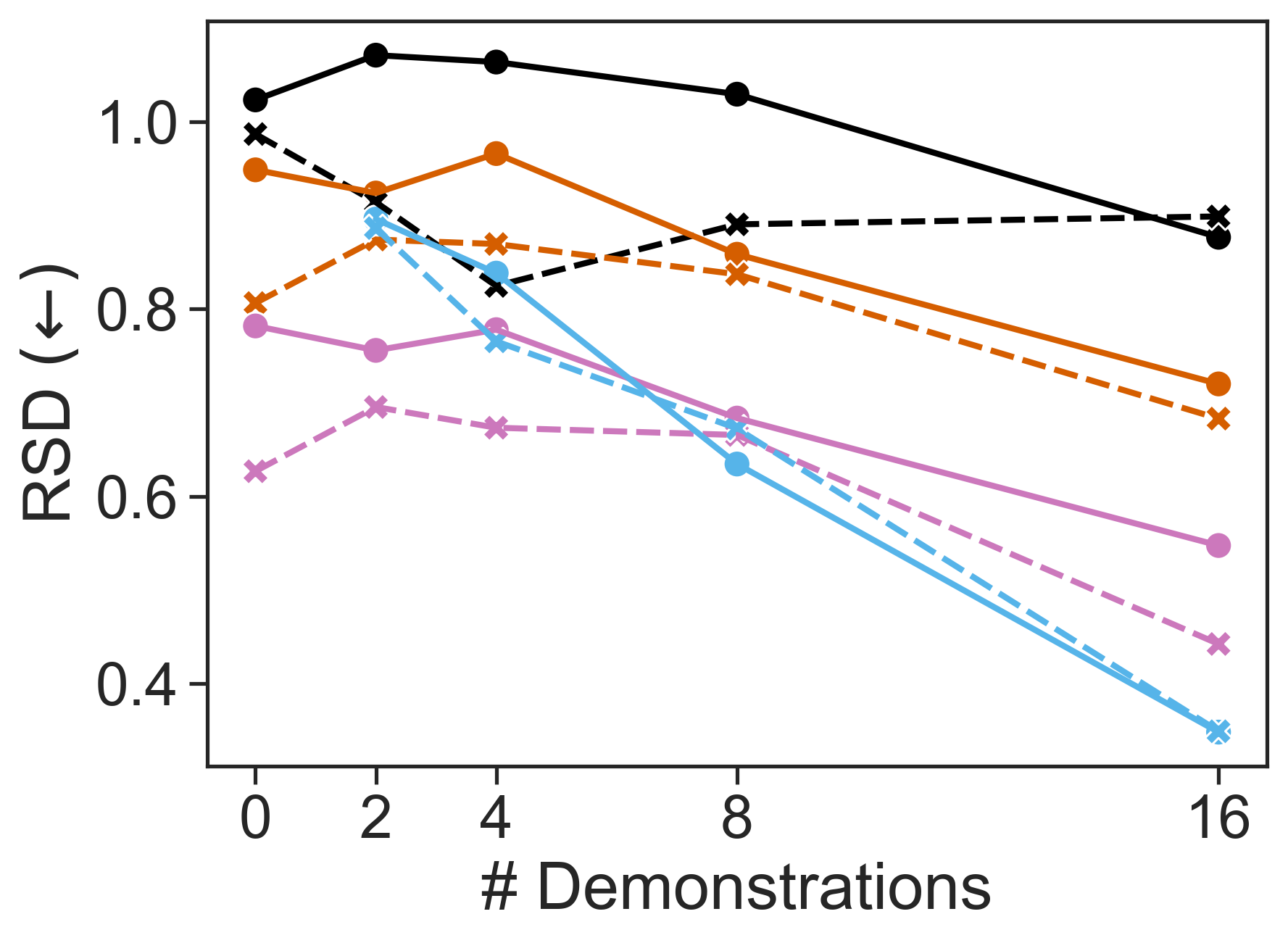}
    \end{subfigure}
    ~
    \begin{subfigure}[b]{0.31\textwidth}
        \centering
                
        \includegraphics[width=\textwidth]{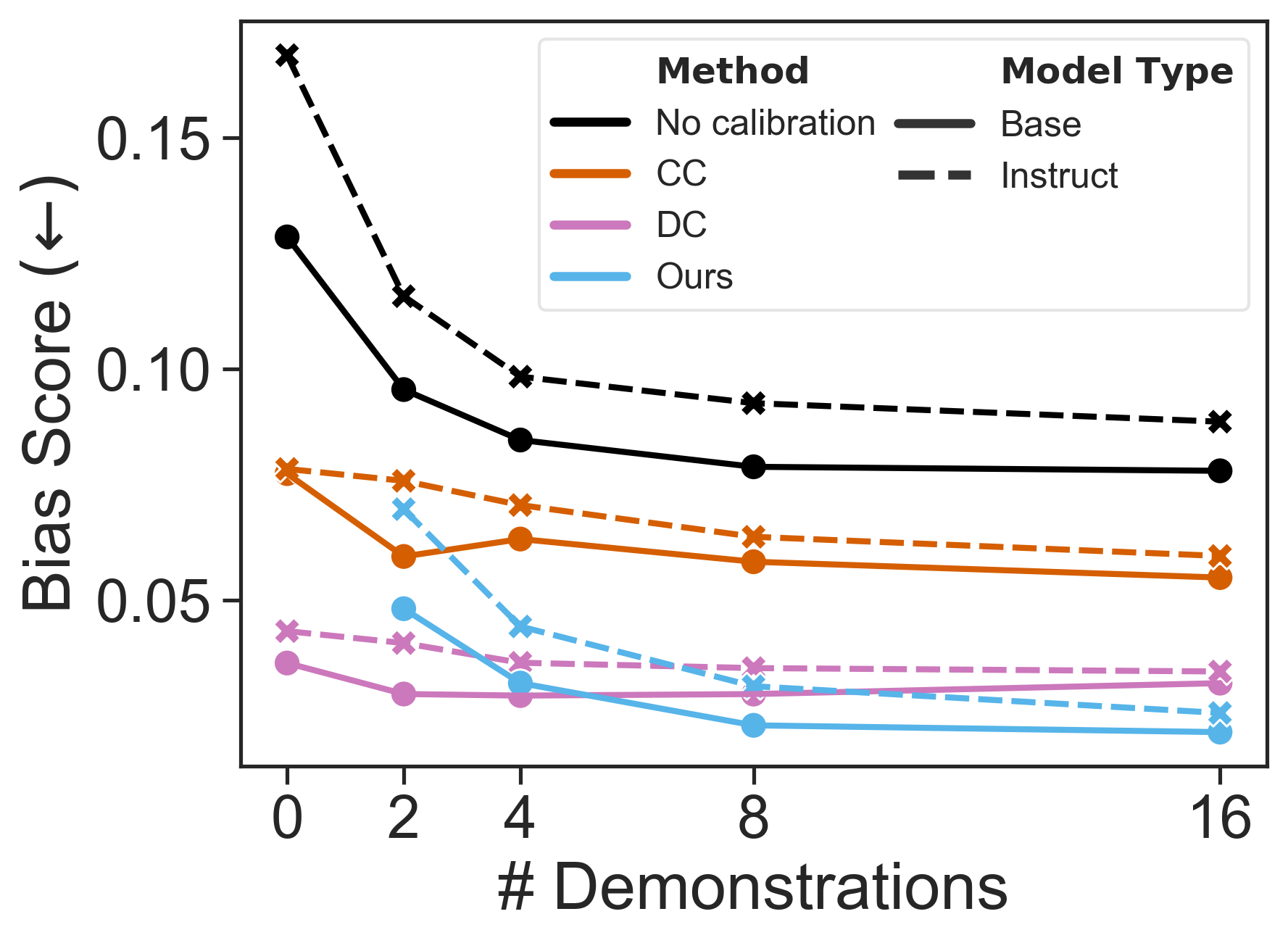}
    \end{subfigure}%

    
    \begin{subfigure}[b]{0.03\textwidth}
        \centering
        \mbox{}
        \vfill
        \rotatebox[origin=c]{90}{\small 40B}
        \bigskip
        \vfill
        \vspace{1.7cm}
        \mbox{}
    \end{subfigure}%
    \begin{subfigure}[b]{0.31\textwidth}
        \centering
        
        \includegraphics[width=\textwidth]{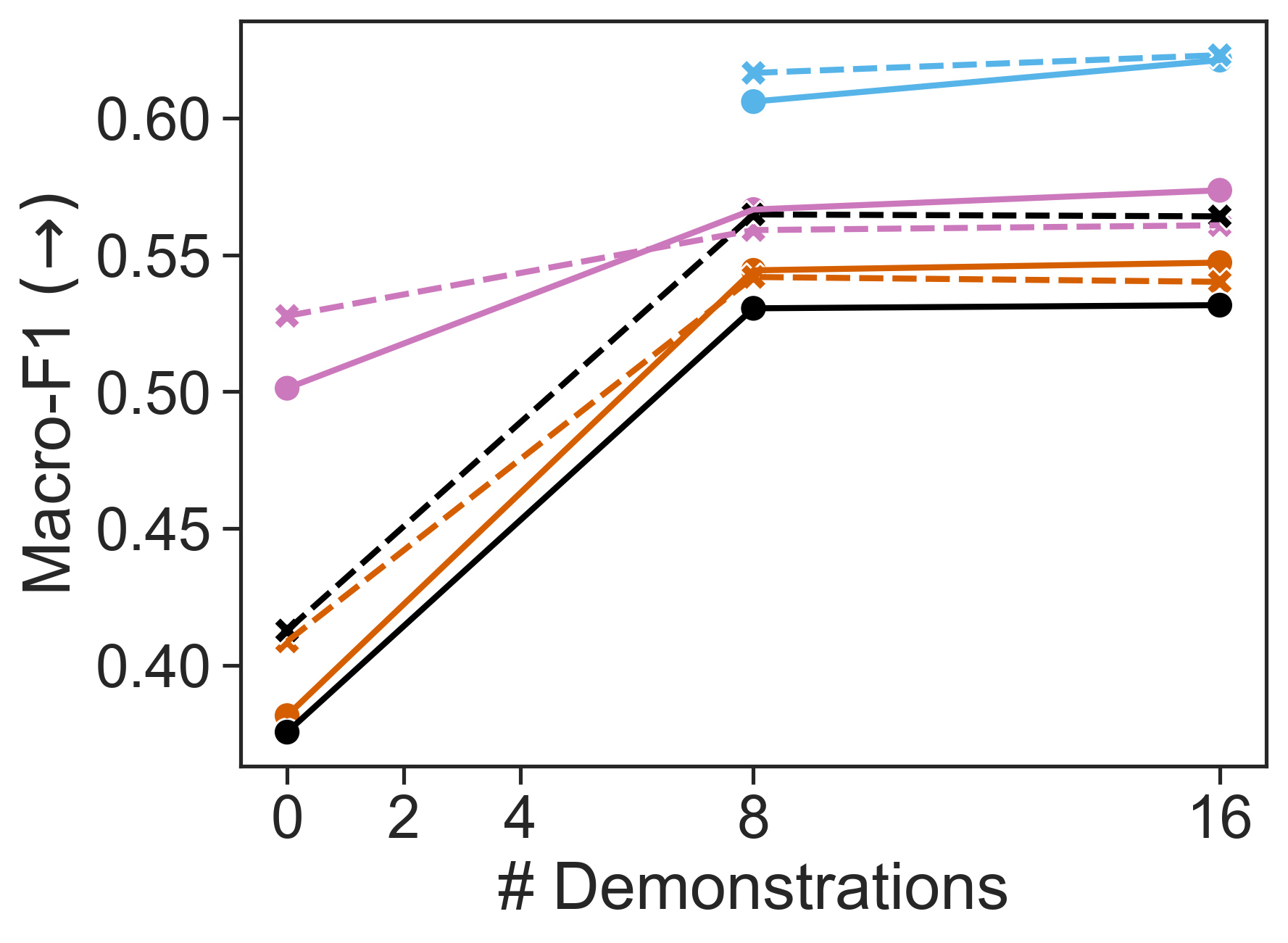}
        
        \subcaption{\hspace{0.5cm}\textsl{Macro-F1}}
        
    \end{subfigure}%
    ~ 
    \begin{subfigure}[b]{0.31\textwidth}
        \centering
        
        \includegraphics[width=\textwidth]{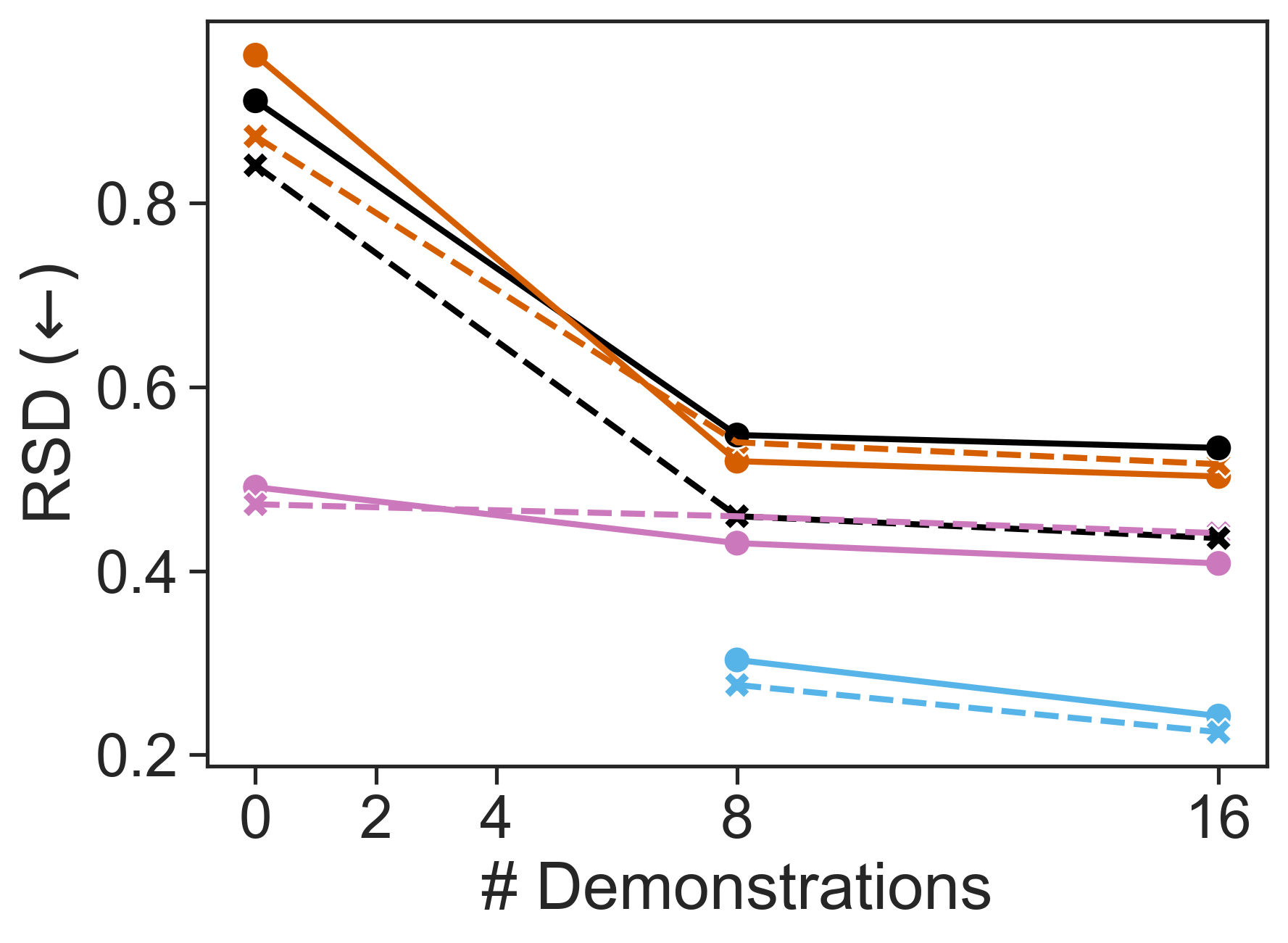}

        \subcaption{\hspace{0.5cm} \textsl{RSD}}

    \end{subfigure}
    ~
    \begin{subfigure}[b]{0.31\textwidth}
        \centering
        
        \includegraphics[width=\textwidth]{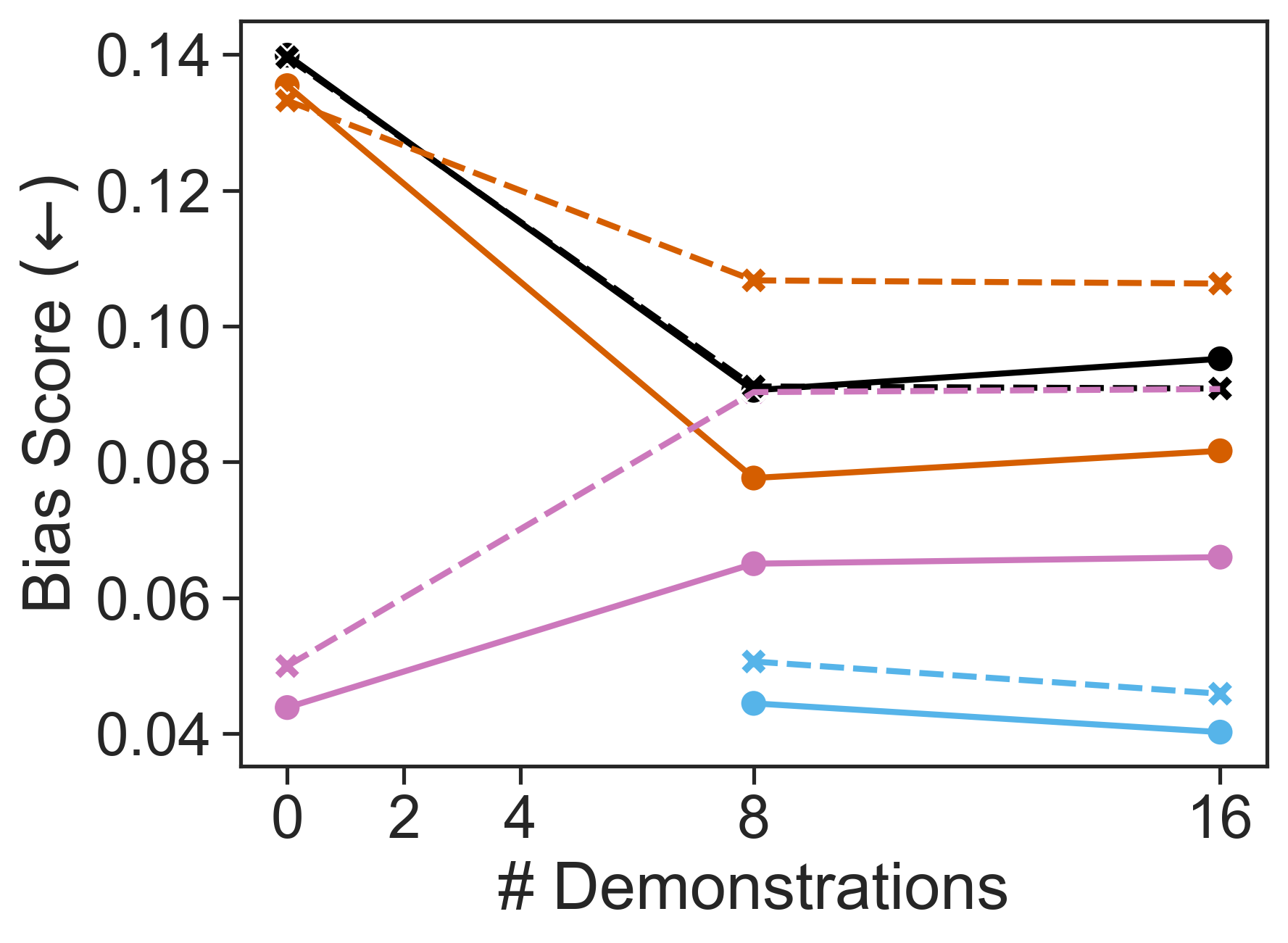}

        \subcaption{\hspace{0.5cm} \textsl{Bias Score}}

    \end{subfigure}%

    \caption{The effect of label bias mitigation methods on performance and bias for Falcon models.}
    
\label{fig:mitigation_methods_falcon}
\end{figure*}